%% file: main.tex
\documentclass[journal]{IEEEtran}
\usepackage{amsmath,amssymb,amsfonts}
\usepackage{cite}
\usepackage{algorithm}
\usepackage{algorithmic}
\usepackage{graphicx}
\usepackage{caption}
\usepackage{lipsum}
\usepackage{enumerate}
\usepackage{amsthm}
\usepackage{color,soul}
\usepackage{booktabs}
\usepackage{threeparttable}
\newcommand{\Exp}{\mathop{\mathbb E}\displaylimits}

\newtheorem{lemma}{Lemma}

\newcommand\VECTOR{}  
\soulregister\cite7 
\newcommand\SPACE{\mathcal}  

\usepackage{makecell}
\ifCLASSOPTIONcompsoc
  \usepackage[caption=false,font=normalsize,labelfont=sf,textfont=sf]{subfig}
\else
  \usepackage[caption=false,font=footnotesize]{subfig}
\fi

\begin{document}

\title{Self-learned Intelligence for Integrated Decision and Control of Automated Vehicles at Signalized Intersections}
\author{Yangang Ren, Jianhua Jiang, Dongjie Yu, Shengbo Eben Li*, Jingliang Duan,  Chen Chen, Keqiang Li
\thanks{This work is supported by Beijing Science and Technology Plan Project with Z191100007419008. It is also partially supported by Tsinghua University-Toyota Joint Research Center for AI Technology of Automated Vehicle. All correspondences should be sent to S. Eben Li with email: lisb04@gmail.com.}
\thanks{Y. Ren, S. Eben Li, D. Yu, C. Chen and K. Li are with State Key Lab of Automotive Safety and Energy, School of Vehicle and Mobility, Tsinghua University, Beijing, 100084, China. They are also with Center for Intelligent Connected Vehicles and Transportation, Tsinghua University. {\tt\small Email: (ryg18, ydj20)@mails.tsinghua.edu.cn; (lishbo, likq)@tsinghua.edu.cn; chenchen2020@mail.tsinghua.edu.cn.}}
\thanks{J. Duan is with the Department of Electrical and Computer Engineering, National University of Singapore, Singapore. {\tt\small Email: duanjl@nus.edu.sg}.
}
\thanks{J. Jiang is with the College of Engineering, China Agricultural University, Beijing, 100084, China. {\tt\small Email: jiangjianhua\_1998@163.com}.
}
}

\maketitle
\begin{abstract}
Intersection is one of the most complex and accident-prone urban scenarios for autonomous driving wherein making safe and computationally efficient decisions is non-trivial.
Current research mainly focuses on the simplified traffic conditions while ignoring the existence of mixed traffic flows, i.e., vehicles, cyclists and pedestrians. 
For urban roads, different participants leads to a quite dynamic and complex interaction, posing great difficulty to learn an intelligent policy. 
This paper develops the dynamic permutation state representation in the framework of integrated decision and control (IDC) to handle signalized intersections with mixed traffic flows.
Specially, this representation introduces an encoding function and summation operator to construct driving states from environmental observation, capable of dealing with different types and variant number of traffic participants. 
A constrained optimal control problem is built wherein the objective involves tracking performance and the constraints for different participants and signal lights are designed respectively to assure safety. We solve this problem by offline optimizing encoding function, value function and policy function, wherein the reasonable state representation will be given by the encoding function and then served as the input of policy and value function.
An off-policy training is designed to reuse observations from driving environment and backpropagation through time is utilized to update the policy function and encoding function jointly.
Verification result shows that the dynamic permutation state representation can enhance the driving performance of IDC, including comfort, decision compliance and safety with a large margin. The trained driving policy can realize efficient and smooth passing in the complex intersection, guaranteeing driving intelligence and safety simultaneously.
\end{abstract}


\IEEEpeerreviewmaketitle

\input{content/1Intro}

\input{content/2Related}

\input{content/3Method}
\input{content/4Simulation}
\input{content/5Results}
\input{content/6Conclusion}



\ifCLASSOPTIONcaptionsoff
  \newpage
\fi

\bibliographystyle{ieeetr}
\bibliography{reference}

\end{document}

%% file: content/1Intro.tex
\section{Introduction}
\IEEEPARstart{I}{nspired} by the success of reinforcement learning (RL) on GO game and robotics \cite{silver2017mastering, mnih2015DQN, gao2019reinforcement}, RL-enabled decision-making has become a promising framework in autonomous driving, which is capable of learning complex policies in high dimensional environments. Intuitively, the goal of RL is to learn policies for sequential decision problems directly by sampling from the
simulator or real experiment and optimizing a cumulative future reward
signal. Unlike supervised learning, RL is viewed as a self-learning algorithm which allows our vehicle agent to optimize its
driving performance by trial-and-error without reliance on manually designed rules and human driving data.
One typical framework to implement RL on driving tasks is the end-to-end pipeline originated by Deep Q-learning (DQN) \cite{mnih2015DQN}, wherein the optimal policy will be trained to output the corresponding control commands with the raw sensor data such as camera images, LiDAR, radar as inputs.
As a pioneering work, Wolf \emph{et al.} (2017) employed the famous DQN to drive a vehicle by choosing the steering wheel angle with the highest Q-value, wherein a $48 \times 27$ pixel gray scale input image is used as the input state \cite{Wolf}. With the 5 discretized actions, they successfully finished the lane-keeping function in a 3D real world physics simulation. After that, Perot \emph{et al.} (2017) adopted the A3C algorithm to  develop the driving policy using only RGB image as input and implemented the continuous longitudinal and lateral control of vehicle in the car racing game\cite{perot2017end}. Besides, other RL algorithms, such as DDPG \cite{lillicrap2015DDPG, wayve}, inverse RL\cite{zou2018inverse}, and SAC\cite{chen2019SACdriving}, were also utilized to construct the policy from simulated images. 
However, these works of RL only focus on simple driving scenarios like path tracking where no surrounding participants are considered because this framework enlarges the complexity of extracting efficient features from observation. For instance, convolutional neural network (CNN) is widely employed to extract critical indicators like speed of ego vehicle or relative distance between vehicles, which is rather difficult for a dense traffic flow\cite{kiran2021deep, dosovitskiy2017carla}.

Besides, some works on RL seeks to accomplish more complex driving tasks concerning the influence of surrounding vehicles. For that purpose, Wang \emph{et al.} (2017) focused on the on-ramp merge scenario and constructed 9 variables in the state representation by considering information of two surrounding vehicles at the target lane\cite{wang2019continuous}. Similarly, Mirchevska \emph{et al.} (2018) built a driving system for highway scenarios, assuming that the ego vehicle could only see the leading and the following vehicles on its own lane and both its adjacent lanes\cite{mirchevska2018high}. 
Duan \emph{et al.} (2020) established a hierarchical RL method for driving where the upper layer generated maneuver selection and the lower layer implemented low-level motion control in both lateral and longitudinal direction. They constructed a 26-dimensional state vector considering 4 nearest surrounding vehicles as well as the road information, destination information \cite{duan2020hierarchical} and the host vehicle successfully learned car-following, lane-changing and overtaking on the two-lane highway.
As for urban scenarios with mixed traffic flows, it seems non-trivial to make a safe and intelligent driving decision with more randomness of bicycles and pedestrians.
Niranjan \emph{et al.} (2020) constructed a rather simple intersection with only 2 pedestrians walking on the zebra stripes and adopted DQN to make the ego vehicle learn to stop \cite{8917299}. Marin \emph{et al.} (2020) developed a collision-avoidance system to assure driving safety at complex scenarios \cite{Toromanoff_2020_CVPR}. They randomly spawn pedestrians crossing the road ahead of the ego vehicle to verify the policy's brake on this situation. Recently, Jiang \emph{et al.} (2021) constructed a more complex intersection scenario with mixed traffic flows and the state design concerns about the 8 vehicles, 4 bicycles and 4 pedestrians nearest to the ego vehicle\cite{jiang2021integrated}. Overall, the state dimension of left-turn, right-turn and straight-going tasks are designed as 64, 32, 36 respectively, based on which three policies are attained by the constraint RL algorithm.
However, these works mainly focus on the stop ability of the ego vehicle w.r.t crossing pedestrians, rather than the entire passing performance at intersections. 

Actually, mixed traffic flows will bring extra difficulties for the successful application of RL algorithms. First and foremost, neural networks are usually adopted as the function approximation of driving policy, which determines that the state of RL-enabled algorithm must be a fixed dimension vector. Therefore, existing research first designed manually a permutation rule for all surrounding vehicles, for example, the relative position to ego vehicle \cite{duan2020hierarchical} or the lane of vehicles \cite{GUAN}, and then choose a fixed number of the closer vehicles as state elements.
However, this formulation is provably not suitable for mixed traffic flows as the number of surrounding traffic participants, as well as their distance order against ego vehicle is constantly changing between two adjacent instants. Especially, pedestrians and bicycles will exacerbate this issue because their walking are not restricted by lanes.
This phenomenon will introduce the discontinuity to RL state, which is prone to decrease the driving performance.
The other issue is the otherness of different traffic participants, saying the learning of driving policy must consider different safety levels for pedestrians, bicycles and vehicles. For instance, the ego vehicle might be reasonable to cut in more frequently w.r.t its surrounding vehicles, while it must tend to wait encountering pedestrians or bicycles because of the high safety priority of vulnerable road users. 

In this paper, we design the dynamic permutation state representation in the framework of integrated decision and control (IDC) to handle the complex intersections with mixed traffic flows. The main contributions and advantages of this paper are summarized as follows:
\begin{enumerate}
\item In order to extract efficient features of mixed traffic flows, we propose dynamic permutation state representation to construct driving states from original environment observation. 
To that end, an encoding network is introduced to project each traffic actor into an unique vector, and the summation of these vectors will be served as the representation of surrounding participants. 
The final driving state will be generated by concatenating the feature of surroundings and that of ego vehicle.
With this encoding function and summation operator, this representation
can handle the varying number of surrounding participants and is permutation invariant to the order of all traffic actors including vehicles, cyclists and pedestrians. Besides, the injectivity of the representation, which says that different observation must correspond to different state, can be guaranteed by carefully designing the architecture of this encoding network.

\item A constrained optimal control problem is developed wherein the objective involves tracking performance within a finite horizon and the constraints aims to assure safety w.r.t. different participants and signal lights. 
We solve this problem offline and output the optimal encoding function, value function and policy function for online application, wherein the reasonable state representation will be given by the encoding function and then served as the input of policy and value function.
After that, policy function will output the front wheel steer and acceleration of ego vehicle, and the value function will predict the tracking performance given one certain path. 
To improve sample utilization, we design an off-policy training to reuse observations from driving environment, from which the finite-horizon tracking performance will be calculated by a predictive manner. 
Then, policy function and encoding function are jointly updated by backpropagation through time, meanwhile the value function is optimized by minimizing the mean square error between its output and the predictive tracking performance.
\end{enumerate}
To verify benefits of the dynamic permutation state representation to IDC, a complex urban intersection scenario is constructed wherein the signal lights, surrounding vehicles, cyclists and pedestrians are considered to construct the original driving observation. Results indicate that the trained policy can realize efficient and smooth passing at this intersection under random traffic flows. And the final policy performance, including comfort, decision compliance and safety, are all improved compared with rule-based methods and original IDC baselines.

The paper is organized as follows. In Section \ref{sec.related_work}, we introduce the key notations and some preliminaries. Section \ref{sec:dpsr} describes dynamic representation state representation and its incorporation with of IDC algorithm. Section \ref{sec:simulation} presents simulation design in complex intersection with mixed traffic flows and Section \ref{sec:result} shows the driving performance comparison. Finally, Section \ref{sec:conclusion} concludes this paper.

%% file: content/2Related.tex
\section{Preliminaries}
\label{sec.related_work}
In this section, we first introduce the principles of reinforcement learning (RL) and the basic idea of integrated decision and control (IDC). Then, current state representation for driving environment will be shown and its disadvantages will be summarized.
\subsection{Basic principle of RL}
Formally, RL aims to seek for the optimal policy $\pi$ which maps from the state space $\mathcal{S}$ to action space $\mathcal{U}$, i.e., $\pi:\mathcal{S}\rightarrow\mathcal{U}$, by the interaction of the agent and its located environment.
At each time step, with a given state $s\in\mathcal{S}$, the agent selects actions $u\in\mathcal{U}$ according to the policy $\pi$, i.e., $u=\pi(s)$, receiving an utility $l(s, u)$ and the next state of the environment $s'=f(s, u)$, where $l:\mathcal{S}\times\mathcal{U}\rightarrow\mathbb{R}$ denotes the utility function and $f:\mathcal{S}\times\mathcal{U}\rightarrow\mathcal{S}$ denotes the system model. 
Value function $v^\pi:\mathcal{S}\rightarrow\mathbb{R}$ is defined as the expected sum of utilities under $\pi$ obtained from the input state $s$, i.e., $v^\pi(s) = \{\sum_{t=0}^{T-1}l_t|s_0=s\}$. Typically, RL algorithm will be composed of two important phases: policy evaluation and policy improvement. 
The former aims to update $v^\pi$ to evaluate the current policy and the latter intends to find another better policy $\pi'$ by minimizing the calculated value function, i.e., $\pi'=\arg\min_{\pi} \mathbb{E}_{s\sim d}\{v^\pi(s)\}$, where $d$ is a state distribution. These two processes will improve the policy progressively until obtaining the optimal counterpart $\pi^{*}$ and its corresponding value $v^{*}$\cite{sutton2018reinforcement}. For practical applications, the policy and value function are usually parameterized as $\pi_{\theta}$ and $V_w$ by neural networks (NNs) to handle continuous and high dimension tasks, in which $\theta$ and $w$ are parameters to be optimized. Under this scheme, many gradient-based optimization methods have been proposed to update these parameters to approximate the optimal policy and value function\cite{schulman2017PPO, schulman2015TRPO,  schulman2017PG_Soft-Q}.
\subsection{Integrated decision and control}
\begin{figure}[!htb]
    \captionsetup{justification=raggedright, 
                  singlelinecheck=false, font=small}
    \centering{\includegraphics[width=0.6\linewidth]{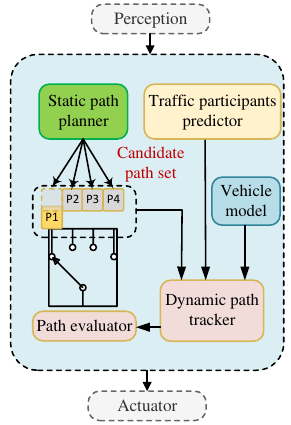}}
    \caption{Integrated decision and control algorithm.}
    \label{fig.IDC}
\end{figure}

As shown in Fig.~\ref{fig.IDC}, integrated decision and control (IDC) mainly consists of static path planner and dynamic path tracker to implement decision-making and control functions\cite{guan2021integrated}. 
Static path planner is used to generate multiple paths only considering static constraints such as road topology, traffic lights. 
Note that these paths will not include time information. Each candidate path is attached with an expected velocity determined by rules from human experience.
Dynamic path tracker aims to select the optimal path and track it considering dynamic obstacles, wherein a finite-horizon constrained optimal control problem is constructed and optimized for each candidate path. The optimal path is selected as the one with the lowest optimal cost function. The IDC framework is computationally efficient because it unloads the heavy online optimizations by solving the constrained problem offline in the form of value function $V_w$  and policy function $\pi_{\theta}$. The policy function is capable of tracking different shape of paths while maintaining the ability to avoid collisions. Meanwhile, the value function can learn to approximate the optimal cost of tracking different paths for the purpose of online path selection. It has been shown in\cite{guan2021integrated} that IDC can output the optimal path and driving actions within 10ms, and is also interpretable in the sense that the solved value and policy functions are the approximation for the optimal cost and the optimal action of the constrained optimal control problem. Moreover, IDC is potential to solve a task-independent problem with tracking errors as objective and safety constraints, making it applicable among a variety of scenarios and tasks.

\subsection{State representation for driving task}
State representation plays a core role for successful implement of RL-enabled algorithms, which aims to design the state $s$ to reasonably describe the driving task based on given observed information. Generally, the observation $\mathcal{O}$ from driving environment should consist of two components: one is the information set of surrounding participants $\mathcal{X}=\{\mathcal{X}^{veh}, \mathcal{X}^{bike}, \mathcal{X}^{ped}\}$, where $\mathcal{X}^{veh}$, $\mathcal{X}^{bike}$, $\mathcal{X}^{ped}$ denote the set of surrounding vehicles, bikes and pedestrians respectively. The other is the feature vector containing other information related to the ego vehicle and road geometry $x_{\rm else} \in \mathbb{R}^{d_2}$, i.e., $\mathcal{O}=\{\mathcal{X},x_{\rm else}\}$.
Furthermore, $\mathcal{X}^{veh}$, $\mathcal{X}^{bike}$, $\mathcal{X}^{ped}$ can be summarized as:
\begin{equation}
\begin{aligned}
\label{eq.surr_participant}
&\mathcal{X}^{veh}=\{x^{veh}_1,x^{veh}_2,...,x^{veh}_L\}, \\
&\mathcal{X}^{bike}=\{x^{bike}_1,x^{bike}_2,...,x^{bike}_M\}, \\
&\mathcal{X}^{ped}=\{x^{ped}_1,x^{ped}_2,...,x^{ped}_N\}, \\
\end{aligned}
\end{equation}
where $x^{veh}_{i} \in \mathbb{R}^{d_1}, x^{bike}_{i} \in \mathbb{R}^{d_1}, x^{ped}_{i} \in \mathbb{R}^{d_1}$ denote the real-valued feature vector of the $i$th vehicle, bike and pedestrian respectively. $L, M$ and $N$ represent the set size of different participants. 
Formally, state is generated by the mapping from observation, i.e., 
\begin{equation}
\begin{aligned}
\label{eq.mapping_s}
s=U(\SPACE{O})=U(\SPACE{X},x_{\rm else}).
\end{aligned}
\end{equation}
One straightforward idea to construct driving state, called fixed permutation state representation, is to directly concatenate the elements in $\mathcal{O}$ together, where the element permutation of in $\mathcal{X}^{veh}, \mathcal{X}^{bike}, \mathcal{X}^{ped}$ is arranged by a predefined sorting rule $o$, for instance, according to distance from ego vehicle,
\begin{equation}
\begin{aligned}
\label{eq.order_mapping}
&s= [U_{\rm F}(\mathcal{X}^{veh}), U_{\rm F}(\mathcal{X}^{bike}), U_{\rm F}(\mathcal{X}^{ped}), x_{\rm else}]^{\top} \\
&s.t. \qquad U_{\rm F}(\mathcal{X}^{veh})=[x_{o(1)}^{veh},\dots,x_{o(L)}^{veh}]^{\top}, \\
& \qquad \quad U_{\rm F}(\mathcal{X}^{bike})=[x_{o(1)}^{bike},\dots,x_{o(M)}^{bike}]^{\top}, \\
& \qquad \quad U_{\rm F}(\mathcal{X}^{ped})=[x_{o(1)}^{ped},\dots,x_{o(N)}^{ped}]^{\top}.
\end{aligned}
\end{equation}
where $U_{\rm F}(\cdot)$ denotes this fixed permutation state representation, which firstly sorts the different participants in terms of distance to the ego vehicle, then concatenating them together to serve as the driving state $s$.

However, for RL-enabled algorithm, this kind of state representation is provably not one good choice as it suffers from the aforementioned permutation sensitivity problem. Firstly, RL algorithm can only consider a fixed number of participants since the input dimension of policy network $\pi_{\theta}$ should be fixed priorly. Therefore, the number of each participant must be reduced or fixed to $L, M, N$ regardless of traffic conditions, leading to the fixed state dimension, i.e., ${\rm dim}(s)=(L+M+N)d_1+d_2$. 
Secondly, different permutations $o$ of $U_{\rm F}(\cdot)$ corresponds to different state vector $s$ and further causes different policy outputs. This phenomenon is exacerbated by the consideration of pedestrians of driving state, wherein their distance order against ego vehicle is constantly changing between two adjacent instants. 

%% file: content/3Method.tex
\section{Integrated decision and control with dynamic permutation state representation}
\label{sec:dpsr}
This section proposes dynamic representation representation algorithm to map the environmental observation to driving states, and incorporates it with IDC to make an intelligent decision at urban scenarios.

\subsection{Dynamic Permutation State Representation}
\begin{figure}[!htb]
    \captionsetup{justification=raggedright, 
                  singlelinecheck=false, font=small}
    \centering{\includegraphics[width=1.0\linewidth]{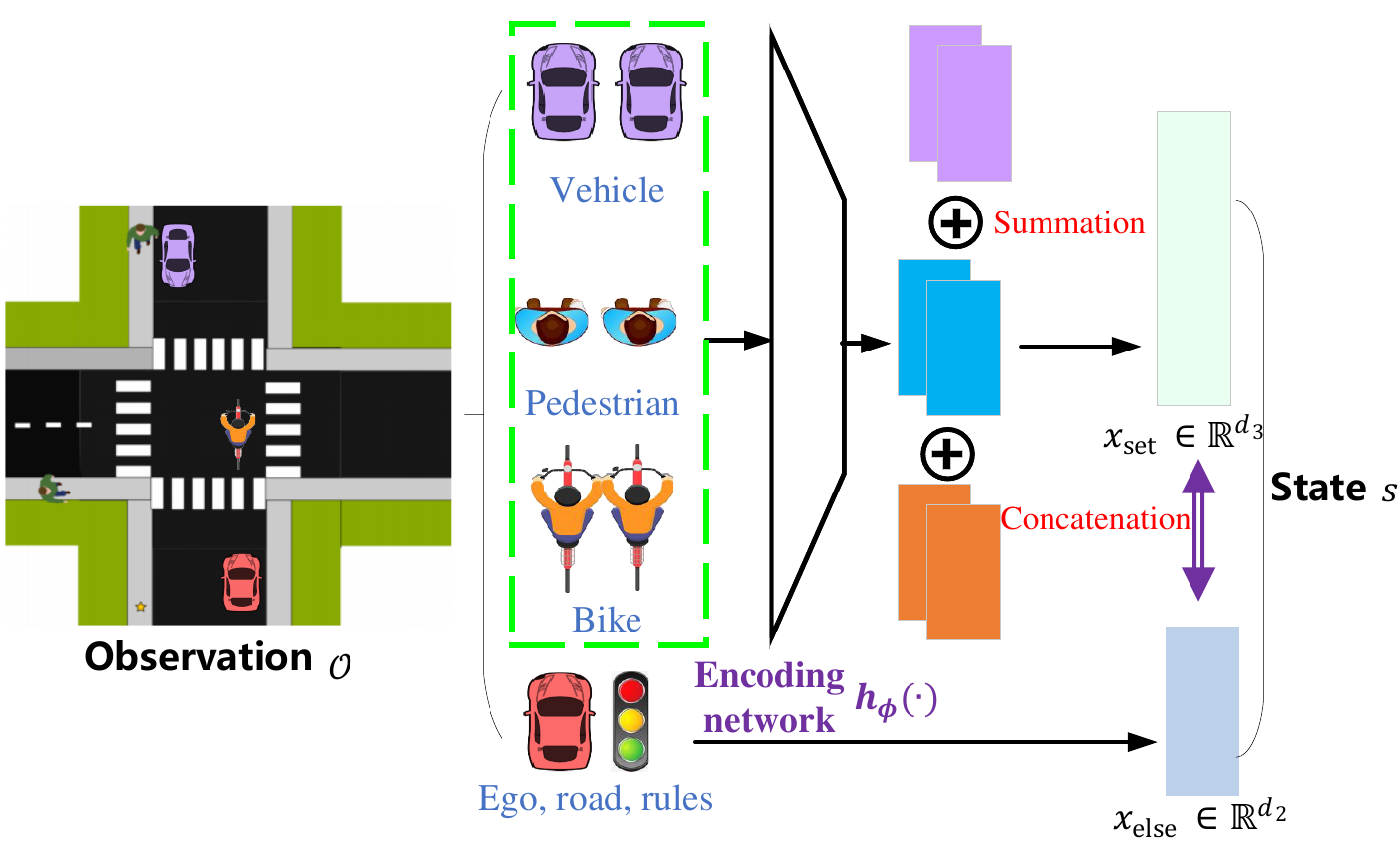}}
    \caption{Dynamic permutation state representation for mixed traffic flows.}
    \label{fig.PI_module}
\end{figure}

We firstly introduce a special encoding network, denoted as $h(x,\phi)$, to extract features of each traffic participant, i.e., vehicles, bikes and pedestrians, as shown in Fig. \eqref{fig.PI_module}.
With the encoding network, each element $x^{veh}_{i}, x^{bike}_{i}, x^{man}_{i}$ can be projected into $h(\VECTOR{x^{veh}_{i}};\VECTOR{\phi}) \in \mathbb{R}^{d_3}, h(\VECTOR{x^{bike}_{i}};\VECTOR{\phi}) \in \mathbb{R}^{d_3}, h(\VECTOR{x^{ped}_{i}};\VECTOR{\phi}) \in \mathbb{R}^{d_3}$ respectively wherein $\phi$ are the parameters of encoding networks. Then a summation operator will be introduced to sum up features of the same participants:
\begin{equation}
\begin{aligned}
\label{eq04:PI_encode}
&\VECTOR{x}_{\rm veh\_{set}}=\sum_{\VECTOR{x^{veh}_{i}}\in\SPACE{X}^{veh}}h(\VECTOR{x^{veh}_{i}};\VECTOR{\phi}), \\
&\VECTOR{x}_{\rm bike\_set}=\sum_{\VECTOR{x^{bike}_{i}}\in\SPACE{X}^{veh}}h(\VECTOR{x^{bike}_{i}};\VECTOR{\phi}), \\
&\VECTOR{x}_{\rm ped\_set}=\sum_{\VECTOR{x^{ped}_{i}}\in\SPACE{X}^{veh}}h(\VECTOR{x^{ped}_{i}};\VECTOR{\phi}),
\end{aligned}
\end{equation}
where $\VECTOR{x}_{\rm veh\_{set}} \in \mathbb{R}^{d_3}, \VECTOR{x}_{\rm bike\_{set}} \in \mathbb{R}^{d_3}, \VECTOR{x}_{\rm ped\_{set}} \in \mathbb{R}^{d_3}$ shows the corresponding encoding vector of surrounding vehicles, bikes and pedestrians. Ultimately, we sum the encoding state of different participants as the features of surrounding participants:
\begin{equation}
\begin{aligned}
\label{eq.sum_state}
&\VECTOR{x}_{\rm {set}}=\VECTOR{x}_{\rm veh\_{set}}+\VECTOR{x}_{\rm bike\_set}+\VECTOR{x}_{\rm ped\_set},\\
\end{aligned}
\end{equation}
and the final driving state can be attained after concatenating $x_{\rm set}$
with other information $\VECTOR{x}_{\rm else}$, i.e.,
\begin{equation}
\begin{aligned}
\label{eq.pi_state}
\VECTOR{s}=U_{\rm D}(\SPACE{O}; \phi) 
=[\VECTOR{x}_{\rm {set}}^\top, \VECTOR{x}_{\rm else}^\top]^\top.
\end{aligned}
\end{equation}
where $U_{\rm D}(\cdot)$ denotes this dynamic permutation state representation. And the state $s$ will be delivered as the input of both policy and value networks to train RL algorithms.

Note that the number of surrounding participants within the perception range of the ego vehicle is constantly changing due to the dynamic nature of the traffic flow. Supposing that the maximum size of three participant sets are $L', M', N'$ respectively, i.e., the set size $L, M, N$ can be varying within $[1, L'], [1, M'], [1, N']$. 
Obviously, the dynamic representation state design in \eqref{eq.pi_state} can deal with dynamic number input of traffic participants. The output encoding vector $\VECTOR{x}_{\rm {set}}$ is always fixed-dimensional and ${\rm dim}(s)=d_2+d_3$ for any combination of $L, M, N$, which can meet the input requirement of neural network.
Secondly, with the summation operator, the encoding vector for each participant in \eqref{eq04:PI_encode} is permutation invariant w.r.t. objects in $\SPACE{X}$. 
More importantly, the injectivity of $U_{\rm D}(\SPACE{O}; \phi)$, which says that different observation $\SPACE{O}$ must generate different state $s$, can be guaranteed by carefully designing the architecture of the feature NN. 
For that, we introduce the space of traffic participants of \eqref{eq.surr_participant}:
\begin{equation}
\begin{aligned}
\nonumber
&\overline{\mathcal{X}^{veh}}=\{\mathcal{X}^{veh}|x_i^{veh}\in\mathbb{R}^{d_1},i\le L,L\in[1,L']\}, \\
&\overline{\mathcal{X}^{bike}}=\{\mathcal{X}^{bike}|x_i^{bike}\in\mathbb{R}^{d_1},i\le M,M\in[1,M']\}, \\
&\overline{\mathcal{X}^{ped}}=\{\mathcal{X}^{ped}|x_i^{ped}\in\mathbb{R}^{d_1},i\le N,N\in[1,N']\}, 
\end{aligned}
\end{equation}
and $\mathcal{X}\in\overline{\mathcal{X}}=\{\overline{\mathcal{X}^{veh}}, \overline{\mathcal{X}^{bike}}, \overline{\mathcal{X}^{ped}}\}$. 
Then we can construct the sufficient condition of injectivity of dynamic representation representation base on the injectivity from \cite{duan2021fixeddimensional}:
\begin{lemma}\label{lemma.encoding}
(Injectivity). Let $\SPACE{O}=\{\SPACE{X},\VECTOR{x}_{\rm else}\}$, where $\VECTOR{x}_{\rm else}\in\mathbb{R}^{d_2}$, $\mathcal{X}\in\overline{\mathcal{X}}$ and $\mathcal{X}=\{\mathcal{X}^{veh}, \mathcal{X}^{bike}, \mathcal{X}^{ped}\}$, $\mathcal{X}^{veh}=\{x^{veh}_1,x^{veh}_2,...,x^{veh}_L\}$,
$\mathcal{X}^{bike}=\{x^{bike}_1,x^{bike}_2,...,x^{bike}_M\}$,
$\mathcal{X}^{ped}=\{x^{ped}_1,x^{ped}_2,...,x^{ped}_N\}$. Denote the maximum set size of $\mathcal{X}^{veh}$, $\mathcal{X}^{bike}$, $\mathcal{X}^{ped}$ are $L', M', N'$ respectively and each item of them $x_i\in\mathbb{R}^{d_1}$ is bounded.
If the feature NN $h(\VECTOR{x};\VECTOR{\phi}):\mathbb{R}^{d_1}\rightarrow\mathbb{R}^{d_3}$ is over-parameterized with a linear output layer, and its output dimension $d_3\ge (L'+M'+N')d_1+1$, there always $\exists \phi^{\dagger}$ such that the mapping $U_{\rm D}(\mathcal{O};\phi^{\dagger}): \overline{\mathcal{X}}\times \mathbb{R}^{d_2}\rightarrow \mathbb{R}^{d_3+d_2}$ in \eqref{eq.pi_state} is injective.
\end{lemma}

\subsection{IDC with dynamic permutation state representation}
Given one path $\tau$ from the candidate path set $\Pi$, i.e., $\tau\in\Pi$, we design a constraint optimal control problem where the objective aims to minimize the expected accumulated utility within a period of predictive horizon, i.e., the cost of path tracking $J_{\rm track}$ w.r.t path $\tau$. And the constraints are designed to assure the safety of ego vehicle against its surrounding participants. The problem formulation can be summarized as follows:
\begin{equation}\label{eq.rl_problem_policy}
\begin{aligned}
\min\limits_{\theta} \quad &J_{\rm track} = \mathbb{E}_{\mathcal{O}_t \sim d}\bigg\{\sum^{T-1}_{i=0}l(s_{i|t}, \pi_{\theta}(s_{i|t}), \tau)\bigg\}\\
{\rm s.t.}\quad & \mathcal{O}_{i+1|t}=f(\mathcal{O}_{i|t}, \pi_{\theta}(s_{i|t}))\\
&s_{i|t}=U_{\rm D}(\SPACE{O}_{i|t}; \phi) \\
&\mathcal{O}_{0|t}=\mathcal{O}_t \\
& g(\mathcal{O}_{i|t})\ge0\\
\end{aligned}
\end{equation}
where $T$ is the prediction horizon and $\mathcal{O}_{i|t}$ is the observation of driving environment at prediction step $i$, starting from the current time step $t$, while $s_{i|t}$ is the corresponding driving state encoded by the dynamic representation state representation.
Note that the initial observation $\mathcal{O}_{0|t}=\mathcal{O}_{t}$ is sampled from real driving environment and $d$ denotes its distribution, which usually is designed as a joint distribution of reference path $\tau$, velocity, speed of ego vehicles and surrounding participants. Then a prediction model $f$ aims to describe the observation transformation within the predictive horizon given the current observation $\mathcal{O}_{i|t}$ and the control policy $\pi_{\theta}(s_{i|t})$. $l(s_{i|t}, \pi_{\theta}(s_{i|t}), \tau)$ denotes the tracking utility concerning the sampled path $\tau$. $g(\mathcal{O}_{i|t})$ denotes all the constraints about the observation $s_{i|t}$, such as the distance to surrounding participants, road edge and stop line under red lights. 
Intuitively, the optimizing process in \eqref{eq.rl_problem_policy} aims to make the driving policy $\pi_{\theta}$ maintain a high-level tracking performance and simultaneously meet the safety requirements.

Obviously, with the existence of the encoding network $h(x,\phi)$,  $\pi_{\theta}(s_{i|t})$ can be rewritten as:
\begin{equation}
\label{eq04:PI_formula}
\begin{aligned}
\pi_{\theta}(s_{i|t})=\pi_{\theta}(U_{\rm D}(\SPACE{O}_{i|t}; \phi)).
\end{aligned}
\end{equation}
which inspires that $h(x,\phi)$ will also contributes to the performance of driving policy.
For such a constraint optimal problem, firstly we employ the penalty function methods \cite{guan2021integrated} to conducted the policy gradient optimization w.r.t the policy and encoding network simultaneously, where it first transforms the constrained problem \eqref{eq.rl_problem_policy} into an unconstrained one, shown as:
\begin{equation}\label{eq.unconstrained_rl}
\begin{aligned}
\min\limits_{\theta, \phi} \quad &J_\pi=J_{\rm track} + \rho J_{\rm safe}\\
&=\mathbb{E}_{\SPACE{O}_{t}}\bigg\{\sum^{T-1}_{i=0}l(s_{i|t}, \pi_{\theta}(s_{i|t}), \tau)\bigg\}+\rho\mathbb{E}_{\SPACE{O}_{t}}\bigg\{\sum^{T-1}_{i=0}\varphi_i(\theta)\bigg\}\\
{\rm s.t.}\quad& \SPACE{O}_{i+1|t}=f(\SPACE{O}_{i|t}, \pi_{\theta}(s_{i|t}))\\
& \SPACE{O}_{0|t}=\SPACE{O}_t \sim d \\
& s_{i|t}=U_{\rm D}(\SPACE{O}_{i|t}; \phi)\\
& \varphi_i(\theta) = \sum[\max\{0, -g(\SPACE{O}_{i|t})\}]^2\\
\end{aligned}
\end{equation}
where $\varphi$ is the penalty function of constraint. Note that the total policy cost $J_\pi$ is composed of the tracking cost $J_{\rm track}$ and safe cost $J_{\rm safe}$, whose importance level is determined by the penalty factor $\rho$.
Then, the policy network will be updated directly to minimize the tracking error and safety cost in \eqref{eq.unconstrained_rl}, i.e., 
\begin{equation}\label{eq.grad_policy}
\begin{aligned}
&\partial_{\theta}J_{\pi}=\partial_{\theta}J_{\rm track} + \rho \partial_{\theta}J_{\rm safe}\\
&=\Exp_{\substack{\SPACE{O}_{0|t} = \SPACE{O}_t \sim d,\\
\SPACE{O}_{i+1|t}=f(\cdot, \cdot), \\s_{i|t}=U_{\rm D}(\SPACE{O}_{i|t}; \phi)
}}\bigg\{\sum^{T-1}_{i=0} \frac{\partial l(s_{i|t}, \pi_{\theta}(s_{i|t}),\tau)}{\partial \theta}+\rho\frac{\partial \varphi_i(\theta)}{\partial \theta}\bigg\}.
\end{aligned}
\end{equation}
Similarly, the encoding network aims to minimize the total policy performance $J_\pi$:
\begin{equation}\label{eq.grad_encoding}
\begin{aligned}
&\partial_{\phi}J_{\pi}=\partial_{\phi}J_{\rm track} + \rho \partial_{\phi}J_{\rm safe}\\
&=\Exp_{\substack{\SPACE{O}_{0|t} = \SPACE{O}_t \sim d,\\
\SPACE{O}_{i+1|t}=f(\cdot, \cdot),\\
s_{i|t}=U_{\rm D}(\SPACE{O}_{i|t}; \phi)}}\bigg\{\sum^{T-1}_{i=0} \frac{\partial l(s_{i|t}, \pi_{\theta}(s_{i|t}),\tau)}{\partial s_{i|t}}\times \frac{\partial s_{i|t}}{\partial \phi } \\
&\qquad\qquad\qquad\qquad\qquad\qquad+\rho\frac{\partial \varphi_i(\theta)}{\partial s_{i|t}}\frac{\partial s_{i|t}}{\partial \phi }\bigg\}.
\end{aligned}
\end{equation}

Besides, the value function $V_w$ is designed to evaluate the tracking cost in terms of different reference paths, for which the training process aims to minimize the error between the output at initial state $s_t$ and the predictive tracking cost $J_{\rm track}$:
\begin{equation}\label{eq.rl_problem_critic}
\begin{aligned}
\min\limits_{w} \quad &J_{V} = \mathbb{E}_{\SPACE{O}_{t}\sim d}\bigg\{\bigg(\sum^{T-1}_{i=0}l(s_{i|t}, \pi_{\theta}(s_{i|t}), \tau) - V_w(s_{t}, \tau)\bigg)^2\bigg\}\\
{\rm where}\quad & \SPACE{O}_{i+1|t}=f(\SPACE{O}_{i|t}, \pi_{\theta}(s_{i|t}))\\
& \SPACE{O}_{0|t} = \SPACE{O}_t\\
& s_{t}=U_{\rm D}(\SPACE{O}_{t}; \phi)
\end{aligned}
\end{equation}
This objective function is an unconstrained optimization problem and we can directly adopt policy gradient to update parameters $w$ of value network:
\begin{equation}\label{eq.grad_value}
\begin{aligned}
\nabla_{w}J_{V}=2\Exp_{\substack{\SPACE{O}_{0|t} = s_t \sim d,\\
\SPACE{O}_{i+1|t}=f(\cdot, \cdot)\\ 
s_{t}=U_{\rm D}(\SPACE{O}_{t}; \phi)}}
\bigg\{\bigg[&V_w(s_{t}, \tau) -\sum^{T-1}_{i=0}l(s_{i|t}, \pi_{\theta}(s_{i|t}), \tau)\bigg] \\ &\times \frac{\partial V_w(s_{t})}{\partial w} \bigg\}.
\end{aligned}
\end{equation}

Accordingly, we not only need to update the parameters of policy and value networks, but also train the encoding network simultaneously in this scheme. Once their optimal counterparts are attained by gradient descent optimization based on \eqref{eq.grad_policy}, \eqref{eq.grad_encoding} and \eqref{eq.grad_value}, these three networks will be implemented online to the driving environment. Specially, the encoding network $h_\phi$ takes charge of mapping the original observation to driving state, which will be served as the input of policy and value network. The value network $V_w$ aims to evaluate tracking performance given a set of candidate paths and choose the optimal one with the lowest cost. After that the optimal policy $\pi_{\theta}$ attempts to track this optimal path meanwhile considering the safety requirements to generate the control command to drive the ego vehicle. 
The training pipeline of IDC with dynamic permutation state representation is shown as Algorithm \ref{alg:RL-DP}.
\begin{algorithm}[!htb]
\caption{IDC with dynamic permutation state representation}
\label{alg:RL-DP}
\begin{algorithmic}
\STATE Initialize parameters $\theta $, $w$, $\phi $
\STATE Initialize learning rate $\beta_\theta, \beta_{w}, \beta_{h}$
\STATE Initialize penalty factor $\rho=1$
\STATE Initialize penalty amplifier $c$, update interval $m$
\STATE Initialize buffer $\mathcal{B}\leftarrow\emptyset$
\STATE Initialize iterative step $k=0$
\REPEAT
\STATE // Sampling (from environment)
      \FOR{each environment step}
      \STATE Receive observation $\SPACE{O}_t$ and calculate state $\VECTOR{s_t}$ using \eqref{eq.pi_state}
      \STATE Add $\VECTOR{s_t}$ into buffer: $\mathcal{B}\cup\{s_t\}$
      \STATE Obtain action $u_t=\pi_{\theta}(s_t)$
      \STATE Apply $u_t$ in environment, returning the next observation $\SPACE{O}_{t+1}$
      \STATE $t=t+1$
      \ENDFOR
\STATE
\STATE // Optimizing
\STATE Fetch a batch of states from $\mathcal{B}$, compute $J_{V}$ and $J_\pi$ by $f(\cdot, \cdot)$, $\pi_{\theta}$ and $h(x,\phi)$
\STATE Update value network with 
\eqref{eq.grad_value}: \\
\qquad \qquad $w  \leftarrow w  - \beta_{w}\nabla_{w }J_{V}$
\STATE Update encoding network with \eqref{eq.grad_encoding}: \\
\qquad \qquad $\phi  \leftarrow \phi  - \beta_{h}\partial_{\phi}J_{\pi}$
\STATE Update policy network with \eqref{eq.grad_policy}: \\
\qquad \qquad $\theta  \leftarrow \theta  - \beta_{\theta}\partial_{\theta} J_{\pi}$
\IF{$k \% m = 0$}
\STATE Update penalty factor $\rho$: \\
\qquad \qquad $\rho\leftarrow c\rho$
\ENDIF
\STATE $k=k+1$
\UNTIL Convergence 
\end{algorithmic}
\end{algorithm}

%% file: content/4Simulation.tex
\section{Implementation}
\label{sec:simulation}
This section constructs an intersection with dense mixed traffic flows and designs the application details to implement IDC on urban driving scenarios.
\subsection{Intersection construction}
\begin{figure}[!htb]
    \captionsetup{justification=raggedright, 
                  singlelinecheck=false, font=small}
    \centering{\includegraphics[width=1.0\linewidth]{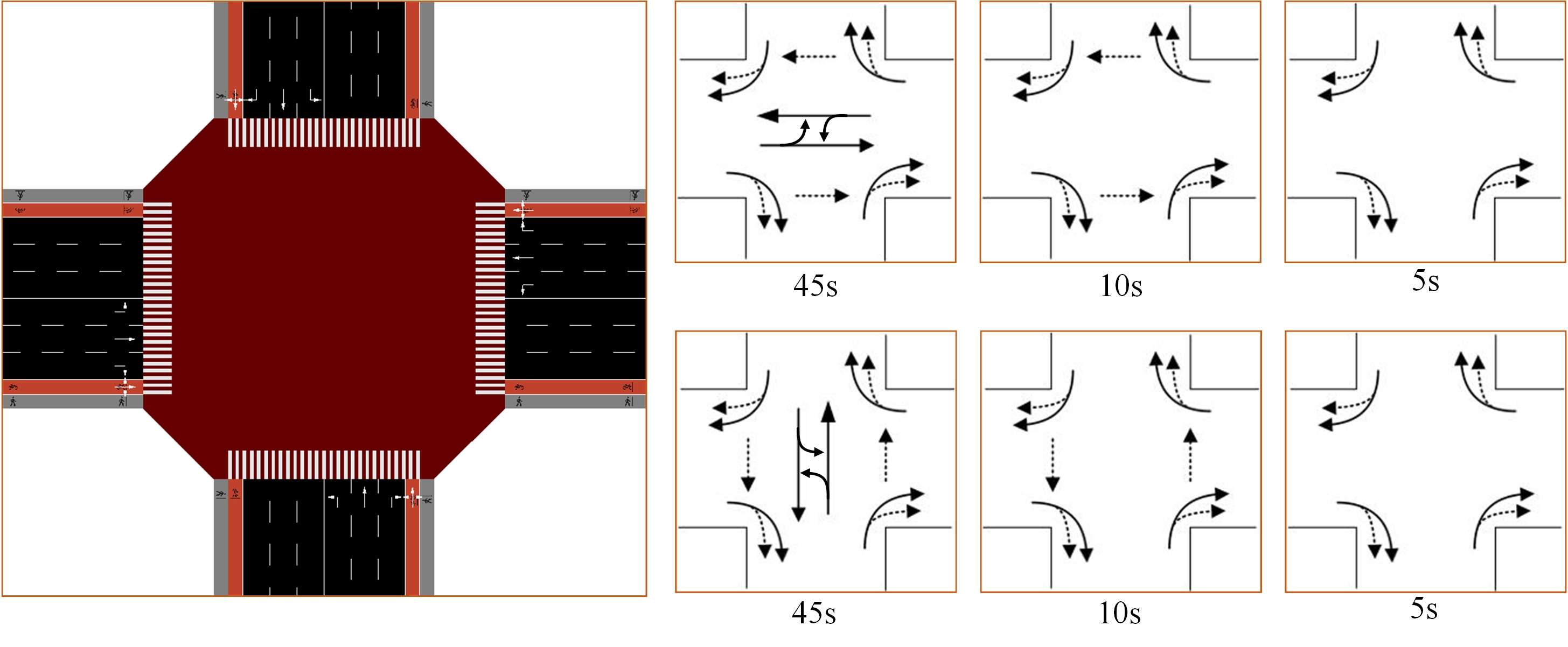}}
    \caption{Intersection structure and traffic light system.}
    \label{fig.intersection}
\end{figure}

As shown in Fig. \eqref{fig.intersection}, we build a signalized four-way intersection equipped with bicycle lanes, sidewalks and crosswalks based on the SUMO software\cite{SUMO2018}. Overall, there are three lanes for motor vehicles, one lane for bicycles and one lane for pedestrians in each driving direction, whose widths are designed as 3.75m, 2.0m, 2.0m respectively. Besides, we generated 400 vehicles, 100 bicycles and 400 pedestrians per hour on each lane, which aims to simulate a dense traffic flow. 
As for traffic light system, a six-phase control system is designed with a cycle time of 120s. Specially,
the signal light controlling the right turn always remains green, and that dominating the left turn and straight going keeps synchronous, which means the traffic flow of straight and turning left will produce more potential conflict points. All surrounding participants are initialized randomly at the beginning of each episode, and their movements are controlled by the embedded car-following and lane-changing models of SUMO. The ego vehicle is initialized outside of the intersection, and aims to complete three different tasks, i.e., turn left, go straight and turn right, to pass this intersection with guaranteeing the driving safety, efficiency and comfort.

\subsection{Observation, utility and action}
\subsubsection{Observation}
Observation should consist of the information of ego vehicle, surrounding traffic participants and road environment perceived by the sensors of ego vehicle. To make a more realistic simulation, we equip the virtual sensor system to the ego vehicle, including camera, radar and lidar. Referring to the specifications of sensor products in market such as Mobileye camera, DELPHI ESR(middle range) and HDL-32E \cite{cao2020novel}, the effective perception ranges of the camera, radar and lidar are set to 80m, 60m and 70m respectively, and the horizontal field of view
of them are set to $\pm 35^\circ$, $\pm 45^\circ$, $360^\circ$ respectively.
Only the surrounding participants within the perception range and not blocked by other participants can be observed. Besides, each variable of surrounding participants is added with noise from a zero-mean Gaussian distribution before being observed. 
The candidate reference paths are generated between center of current lane and the potential target lane, resulting in three candidate paths for each task.

The concrete variables of observation are listed in Table \ref{tab:observation}.
For each surrounding participant like the vehicle, bike and pedestrian, 
we consider the longitudinal and lateral position $p_{x}^{\rm other}$ and  $p_{y}^{\rm other}$, speed $v_{\rm other}$, heading angle $\Phi_{\rm other}$, length $L_{\rm other}$, width $W_{\rm other}$ and participant type $K$, i.e., $\VECTOR{x}=[p_{x}^{\rm other},p_{y}^{\rm other},v_{\rm other},\Phi_{\rm other},L_{\rm other},W_{\rm other}, K]^\top$. Note that $K$ is designed as a special variable to indicate different participants, i.e., $K=0, 1, 2$ represent vehicle type, bike type and pedestrian type respectively.
In addition, $x_{\rm else}$ is designed as a 24-dimensional vector, which contains 8 indicators for the ego vehicle, 1 indicator for the traffic light phase, 3 indicators for the tracking error and 12 indicators for the reference path information. Specifically, the information of the ego vehicle consists of the longitudinal coordinate $p_{\rm x}^{\rm ego}$, lateral coordinate $p_{\rm y}^{\rm ego}$, longitudinal speed $v_{x}$, lateral speed $v_{y}$, heading angle $\Phi$, yaw rate $\omega$, length $L_{\rm ego}$ and width $W_{\rm ego}$. The status of the traffic light $l_{phase}$ is indicated by the index of phase, which belongs to ${\{0, 1, 2, 3, 4, 5\}}$. Tracking error is constructed by the current ego vehicle position and its corresponding reference point, including distance error $\Delta{p}$, speed error $\Delta{v}$ and heading error $\Delta\Phi$. Furthermore, the reference path information at 5, 10 and 15 meters ahead the ego vehicle are utilized to describe the path shape, wherein each point contains horizontal coordinate ${x_{\rm ref}}$, vertical coordinate ${y_{\rm ref}}$ and heading angle ${\Phi_{\rm ref}}$ and expected velocity ${v_{\rm ref}}$. We should emphasize that the position information of surrounding participants are constructed as the relative distance against ego vehicle. i.e., $p_{x}^{\rm other}-p_{x}^{\rm ego}, p_{y}^{\rm other}-p_{y}^{\rm ego}$ to reflect their interactions.
See Table \ref{tab:observation} for more specific details.
\begin{table}[!htb]
\centering
\caption{Observation design}
\label{tab:observation}
\begin{tabular}{cccc}
\toprule
$\mathcal{O}$&Name & Symbol &Unit \\
\midrule
$x_{\rm veh},$&Relative longitudinal position&$p_{x}^{\rm other}-p_{x}^{\rm ego}$&m\\
$x_{\rm bike},$&Relative lateral position&$p_{y}^{\rm other}-p_{y}^{\rm ego}$&m\\
$x_{\rm ped}$&Speed&$v_{\rm other}$&m/s\\
&Heading angle&$\Phi_{\rm other}$& rad \\
&Length&$L_{\rm other}$& m \\
&Width&$W_{\rm other}$& m \\
&Type &$K$&- \\
\midrule
$x_{\rm else}$&Longitudinal position of ego vehicle&$p_{x}^{\rm ego}$&m\\
&Lateral position of ego vehicle&$p_{y}^{\rm ego}$&m\\
&Longitudinal speed of ego vehicle&$v_{x}$&m/s\\
&Lateral speed of ego vehicle&$v_{y}$&m/s\\
&Heading angle of ego vehicle&$\Phi$& rad \\
&Yaw rate of ego vehicle&$\omega$& rad/s \\
&Length of ego vehicle&$L_{\rm ego}$& m \\
&Width of ego vehicle&$W_{\rm ego}$& m \\
&Traffic light&$l_{phase}$&-\\
&Longitudinal position of ahead reference &${x_{\rm ref}}$&{m}\\
&Lateral position of ahead reference&${y_{\rm ref}}$&{m}\\
&Heading angle of ahead reference &${\Phi_{\rm ref}}$&{rad}\\
&Velocity of ahead reference &${v_{\rm ref}}$&{m/s}\\
&Distance error&$\Delta{p}$&m\\
&Speed error&$\Delta{v}$&m/s\\
&Heading angle error&$\Delta\Phi$& rad \\
\bottomrule
\end{tabular}
\end{table}

\subsubsection{Action Design}
We utilize classic dynamic bicycle model for ego vehicle \cite{ge2020numerically} and choose the front wheel angle and expected acceleration, denoted as $\delta$, $a$, to realize the longitudinal and lateral control, i.e., $u=[\delta, a]^\top$.
Considering the vehicle actuator saturation, the action execution shall be limited to a certain range. Hence, we assume $\delta\in[-0.4,0.4]$ rad, $a\in[-3.0,1.5]$ m/${\rm{s}}^2$. 

\subsubsection{Utility}
As utility $l(\cdot,\cdot)$ mainly involves in the precision, stability and energy-saving of path tracking, here we choose the tracking error including $\Delta{p}$, $\Delta{v}$ and $\Delta\Phi$, yaw rate of ego vehicle $\omega$ and control actions $u$ to construct a classic quadratic form utility:
\begin{equation}
\label{eq:reward}
\nonumber
\begin{aligned}
l(\cdot,\cdot)=0.05{\Delta{v}}^2+&0.8{\Delta{p}}^2+30{\Delta{\phi}}^2+0.02{\omega}^2 \\
&+2.5{\delta}^2+2.5{\dot\delta}^2+0.05{a}^2+0.05{\dot{a}}^2.
\end{aligned}
\end{equation}
where $\dot\delta$ and $\dot{a}$ denotes the derivation of steering wheel and acceleration, leading the driving policy to output smooth control commands.
\subsection{Constraint design}
\begin{figure}[!htb]
    \captionsetup{justification=raggedright, 
                  singlelinecheck=false, font=small}
    \centering{\includegraphics[width=0.9\linewidth]{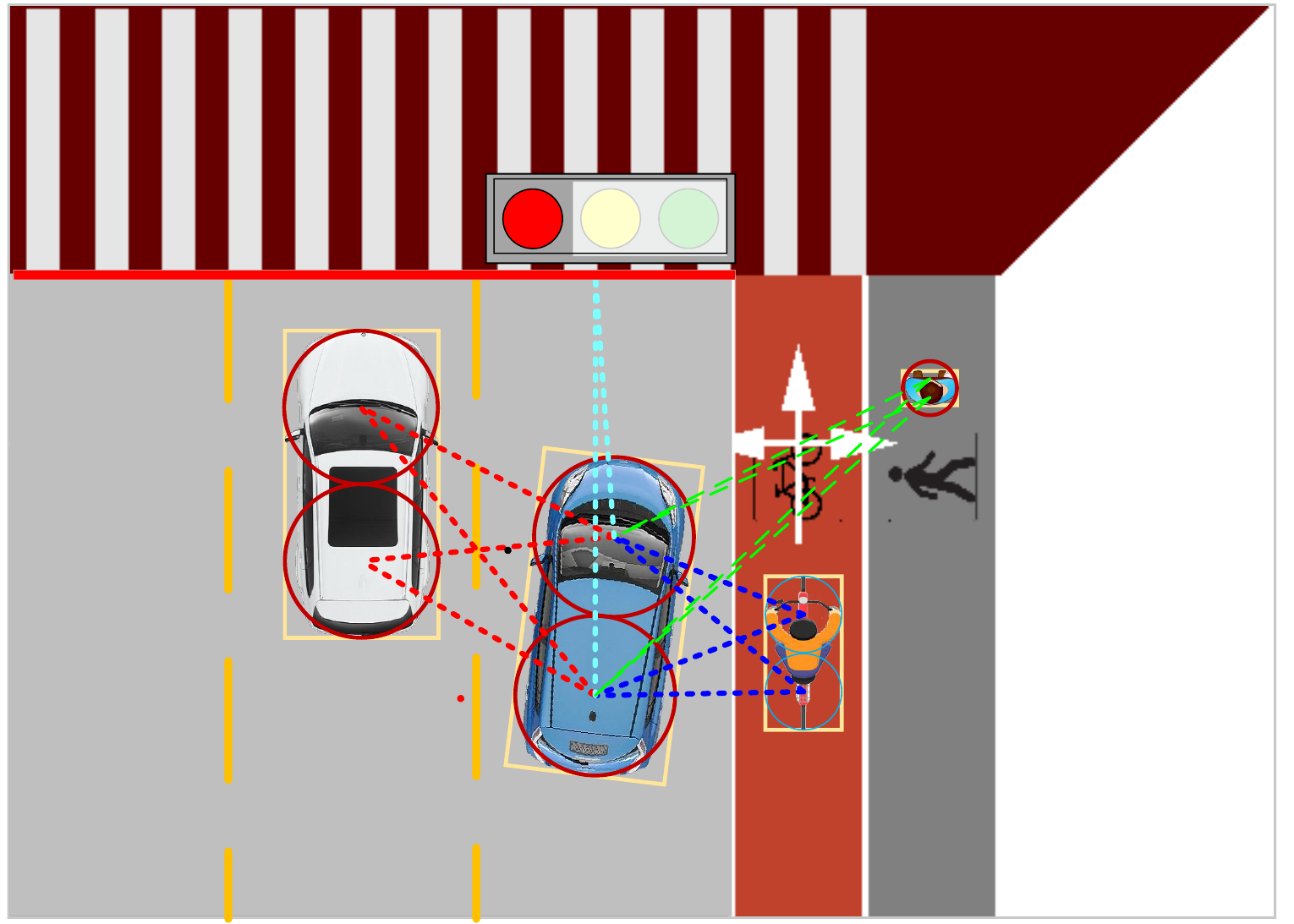}}
    \caption{Safety constraint design.}
    \label{fig.constraints}
\end{figure}

Constraints are crucial to assure the driving safety in IDC. Here we represent each dynamic participant, including the vehicles, bikes and pedestrians by two circles as illustrated by Fig. \ref{fig.constraints}, wherein the front and rear center are determined by the current position, shape and heading angle. Take one participant as an example, its front center $O_{\rm other}^{F}$ and rear center $O_{\rm other}^{R}$ can be calculated as:
\begin{equation}
\label{eq:reward}
\nonumber
\begin{aligned}
O_{\rm other}^{F}=
\begin{bmatrix}
p_{x}^{\rm other}+\frac{L_{\rm other}+W_{\rm other}}{2}\rm{cos}\Phi_{\rm other}\\
p_{y}^{\rm other}+\frac{L_{\rm other}+W_{\rm other}}{2}\rm{sin}\Phi_{\rm other}\\
\end{bmatrix},\\
O_{\rm other}^{R}=
\begin{bmatrix}
p_{x}^{\rm other}-\frac{L_{\rm other}+W_{\rm other}}{2}\rm{cos}\Phi_{\rm other}\\
p_{y}^{\rm other}-\frac{L_{\rm other}+W_{\rm other}}{2}\rm{sin}\Phi_{\rm other}\\
\end{bmatrix}.
\end{aligned}
\end{equation}
Similarly, the ego vehicle also possesses corresponding front center $O_{\rm ego}^{F}$ and rear center $O_{\rm ego}^{R}$:
\begin{equation}
\label{eq:reward}
\nonumber
\begin{aligned}
O_{\rm ego}^{F}=
\begin{bmatrix}
p_{x}^{\rm ego}+\frac{L_{\rm ego}+W_{\rm ego}}{2}\rm{cos}\Phi\\
p_{y}^{\rm ego}+\frac{L_{\rm ego}+W_{\rm ego}}{2}\rm{sin}\Phi\\
\end{bmatrix},\\
O_{\rm ego}^{R}=
\begin{bmatrix}
p_{x}^{\rm ego}-\frac{L_{\rm ego}+W_{\rm ego}}{2}\rm{cos}\Phi\\
p_{y}^{\rm ego}-\frac{L_{\rm ego}+W_{\rm ego}}{2}\rm{sin}\Phi\\
\end{bmatrix},
\end{aligned}
\end{equation}
And a typical constraint based on distance comparison can be defined as:
\begin{equation}
\nonumber
\label{eq:reward}
d(O_{\rm ego}^{\dagger}, O_{\rm other}^{\ast})\ge r_{\rm other} + r_{\rm ego}, \ast\in \{F,R\}, \dagger\in \{F,R\}
\end{equation}
where $r_{\rm other}$ and $r_{\rm ego}$ are radii of circles of surrounding participant and the ego vehicle. Obviously, four constraints are constructed between each participant and the ego vehicle, and safety is thought to be satisfied if the distance of two points is more than the radii sum of these two participants.
\begin{table}[h]
\centering
\caption{Parameters of constraints}
\begin{tabular}{lccc}
\hline
                    &Vehicle    &Bicycle    & Pedestrian \\ \hline
$r_{\rm other}$[m]  & 1.75 & 2.0  & 2.2        \\
$L_{\rm other}$[m]       & 4.8         & 2.0          & 0.48          \\
$W_{\rm other}$[m]    & 2.0 & 0.48  & 0.48\\ \hline
\end{tabular}
\label{table:constraint}
\end{table}
Considering the distinct property of surrounding participants w.r.t the shape and safety priority, we design different safety radii for them, as listed in Table \ref{table:constraint}. 
Note that $r_{\rm ego}=1.75$m and $r_{\rm other}$ is designed as 1.75m, 2.0m and 2.2m respectively for surrounding vehicles, bikes and pedestrians such that the vulnerable road users have the highest protection priority. Additionally, the light constraint is added by the distance of ego vehicle to the stop line, denoted as $O_{SL}=[SL_x, SL_y]^\top$ where $SL_x, SL_y$ represents the longitudinal and lateral position of stop line center. Concerning the ego vehicle locates outside of intersection and faces a red light in left-turn or straight-going task, the constraints on red lights will make a difference:
\begin{equation}
\nonumber
\label{eq:reward}
d(O_{\rm ego}^{\dagger}, O_{SL})\ge D_{SL}, \dagger\in \{F,R\}
\end{equation}
where $D_{SL}=0.5m$ is the safe distance to stop line. 

%% file: content/5Results.tex
\section{Simulation}
With the constructed scenario, here we conduct the training pipeline under the scheme of IDC and implement the trained three network functions online to verify their driving performance.
\label{sec:result}
\subsection{Comparison of training}
Firstly, we compare the training performance of IDC with dynamic permutation state representation and original IDC, i.e., IDC with fixed permutation state representation, wherein the only difference lies in taking $s=U_{\rm F}(\SPACE{O})$ or $s=U_{\rm D}(\SPACE{O})$. For fixed permutation state representation, it must adopt $U_{\rm F}(\SPACE{O})$ in \eqref{eq.order_mapping} to construct driving state for the missing of encoding network. Thus, the number of surrounding traffic participants must be fixed priorly and the nearest 8 vehicles, 4 bicycles and 4 pedestrians are considered to construct driving states, each type of which are sorted in increasing order according to relative distance to ego vehicle. 
For dynamic permutation state representation, we can consider all observed surrounding traffic participants within the sensor range, i.e., $L\in[1,L'], M\in[1,M'], N\in[1,N']$ are constantly changing. And the maximum number $L', M', N'$ are set to 10, 6, 6 respectively. According to Lemma \ref{lemma.encoding}, the output dimension $d_3$ should satisfy that $d_3\ge (L'+M'+N')d_1+1=155$. Hence, we assume $d_3=155$. The networks of policy, value and encoding function employ similar architecture, which contains 2 hidden layers, with 256 units per layer. All hidden layers take Gaussian Error Linear Units (GELU)  \cite{hendrycks2016gelu} as activation functions. The Adam method \cite{Diederik2015Adam} with a cosine annealing learning rate is adopted to update all networks. The predictive horizon $T$ is set to be 25, which is 2.5s in practice. See Table \ref{tab.hyper} for more details.
\begin{table}[!htp] 
\centering
\caption{Training hyperparameters}
\label{tab.hyper}
\begin{threeparttable}[h]
\begin{tabular}{lc}
\toprule
Hyperparameters & Value \\
\hline
\quad Optimizer &  Adam ($\beta_{1}=0.9, \beta_{2}=0.999$)\\
\quad Approximation function  & MLP \\
\quad Number of hidden layers & 2\\
\quad Number of hidden units & 256\\
\quad Nonlinearity of hidden layer& GELU\\
\quad Replay buffer size & 5e5\\
\quad Batch size & 256\\
\quad Policy learning rate & cosine annealing 3e-4 $\rightarrow$ 1e-5 \\
\quad Value learning rate & cosine annealing 8e-4 $\rightarrow$ 1e-5\\
\quad Encoding learning rate & cosine annealing 8e-4 $\rightarrow$ 1e-5\\
\quad Penalty amplifier $c$ & 1.1\\
\quad Total iteration & 200000 \\
\quad Update interval $m$ & 100 \\
\quad Number of Actors &4\\
\quad Number of Buffers &4\\
\quad Number of Learners &8\\
\bottomrule
\end{tabular}
\end{threeparttable}
\end{table}

\begin{figure}[!htb]
\captionsetup[subfigure]{justification=centering}
\subfloat[Total policy cost]{\includegraphics[width=0.5\linewidth]{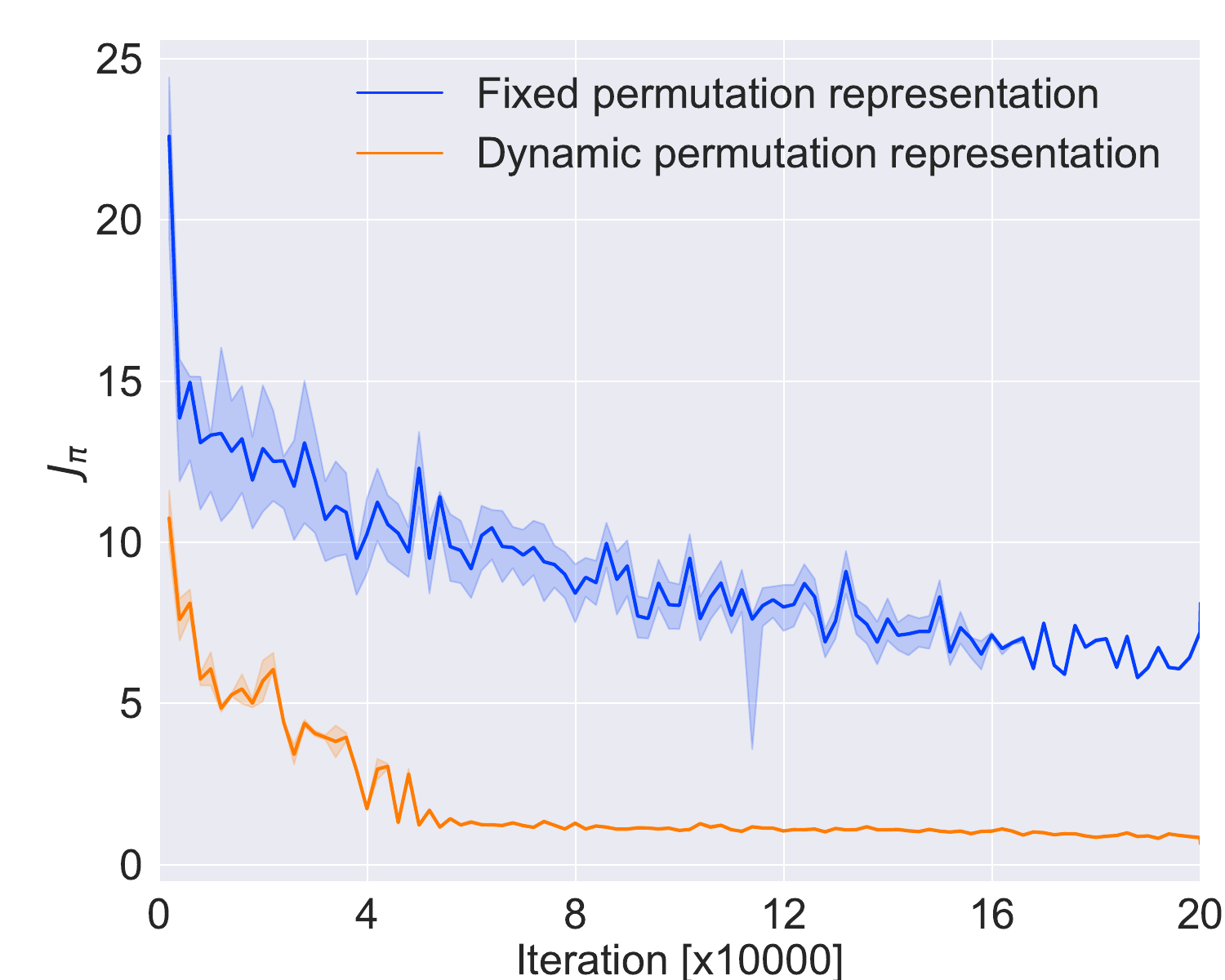}}
\subfloat[Value cost]{\includegraphics[width=0.5\linewidth]{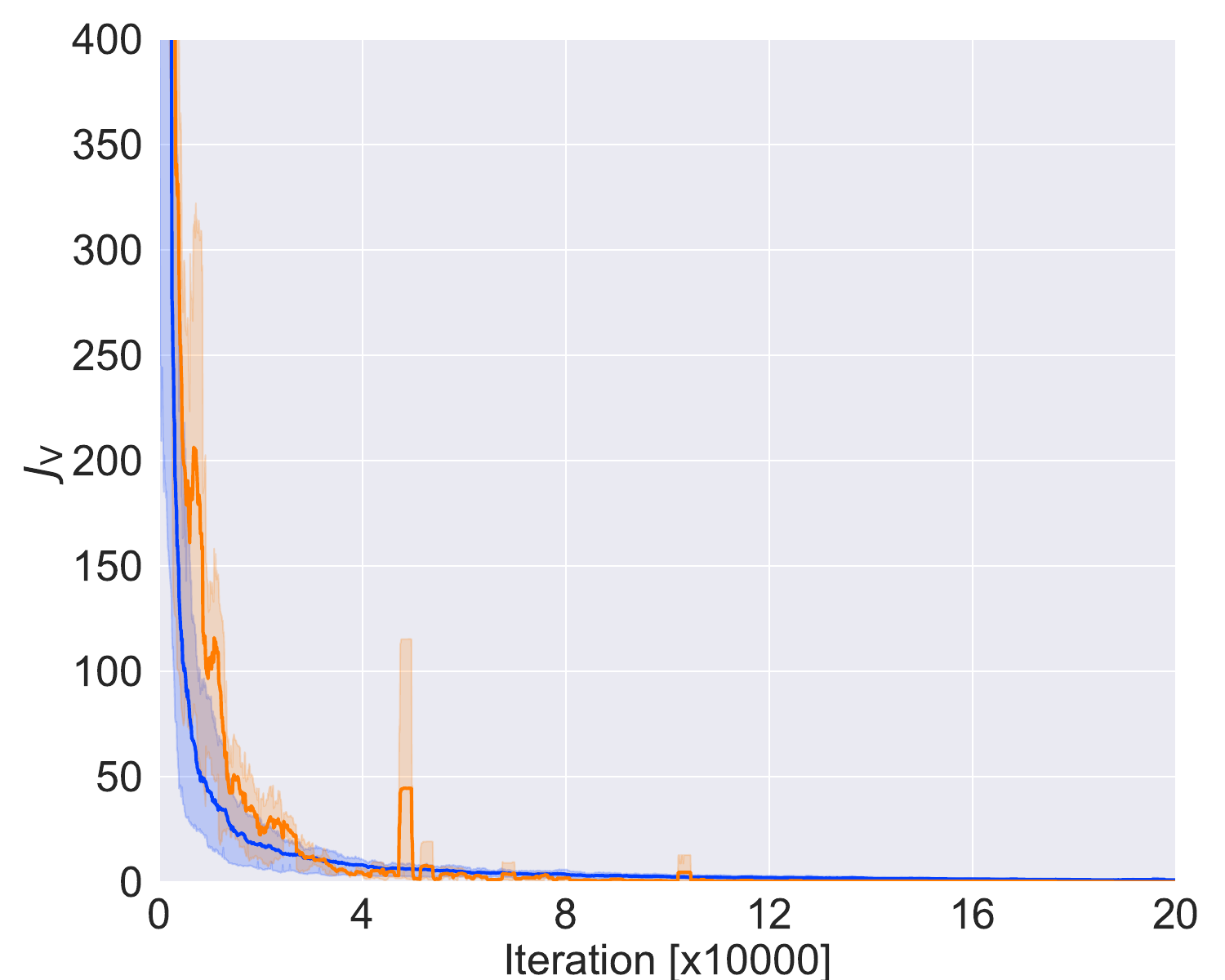}}\\
\subfloat[Tracking cost of policy]{\includegraphics[width=0.5\linewidth]{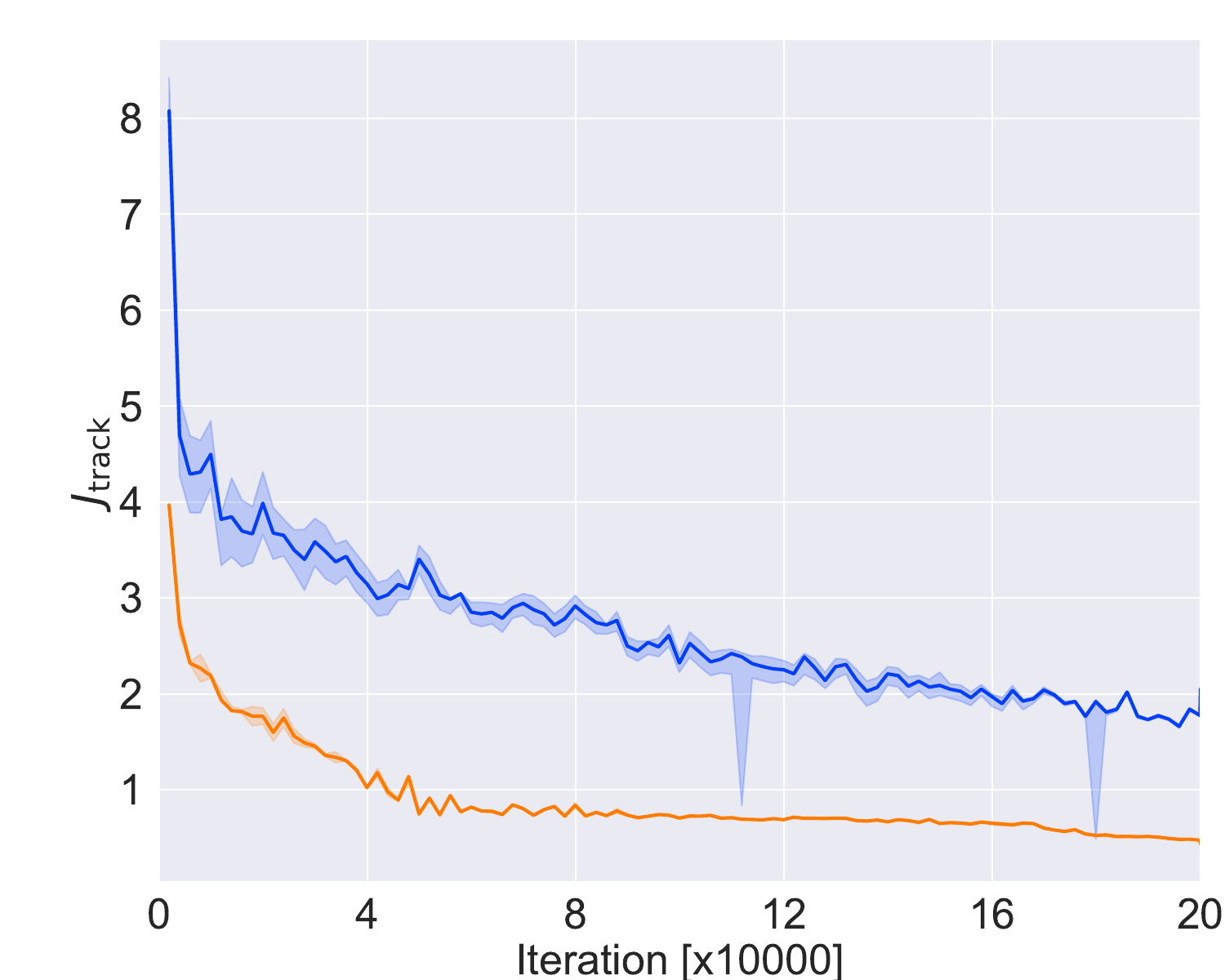}}
\subfloat[Safety cost of policy]{\includegraphics[width=0.5\linewidth]{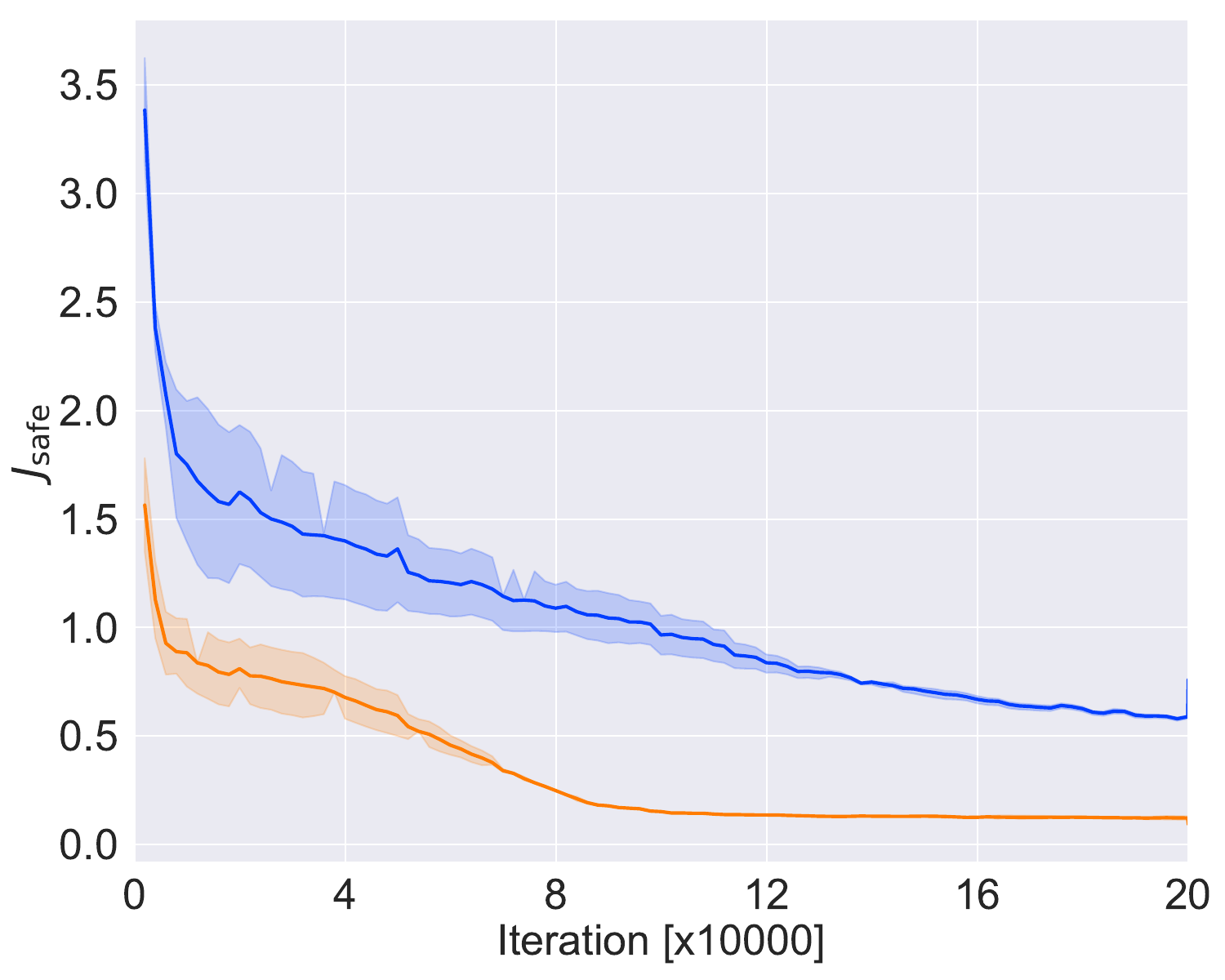}}\\
\caption{Training performance of IDC. The solid lines correspond to the mean and the shaded regions correspond to 95\% confidence interval over 5 runs.}
\label{f:return}
\end{figure}

During training process, we record the policy performance and value loss every 1000 iterations and Fig.~\ref{f:return} demonstrates the learning curves of IDC combined with different state representation. We can see from Fig.~\ref{f:return}(a) that the total policy cost $J_\pi$ of dynamic representation presents a significant reduction compared with that of fixed representation, indicating a better policy has been obtained. Furthermore, we monitor the components of $J_\pi$, i.e., the tracking cost $J_{\rm track}$ and safety cost $J_{\rm safe}$ respectively in Fig.~\ref{f:return}(c) and (d). Both of them have demonstrated a decreasing tendency, meaning that the encoding network in IDC indeed makes a difference in the training of policy, boosting the tracking performance and safety requirement jointly. 
Meanwhile, the value cost in Fig.~\ref{f:return}(b) of these two methods shows more similar decreases to 0 because the learning seems to be much easier as a paradigm of supervised learning. However, the policy learning relies on the interaction with driving environment, thus extracting efficient features will be beneficial to driving performance.

\begin{figure*}[!htbp]
\centering
\captionsetup[subfigure]{justification=centering}
\subfloat[t=0.0s]{\includegraphics[width=0.25\linewidth]{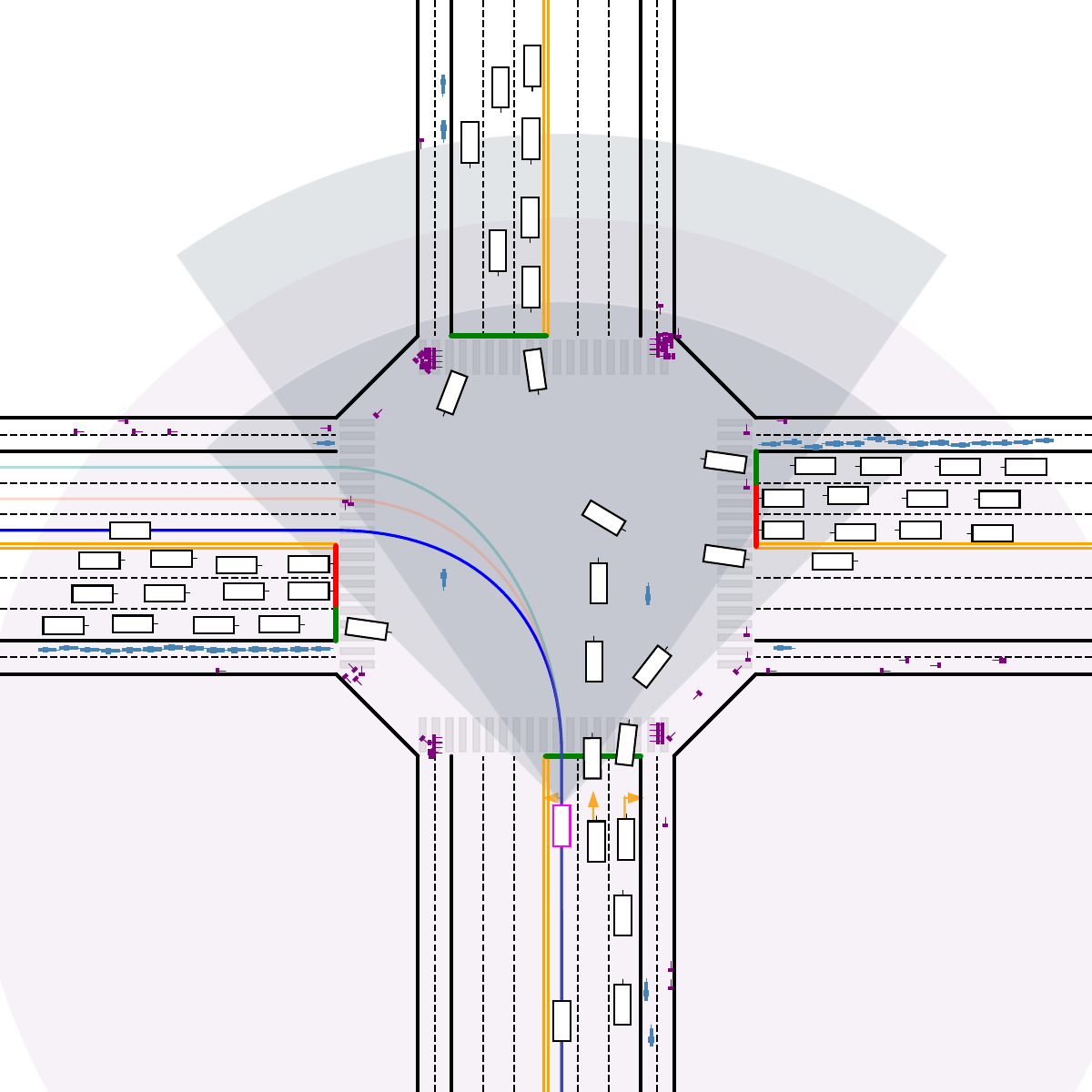}}
\subfloat[t=8.0s]{\includegraphics[width=0.25\linewidth]{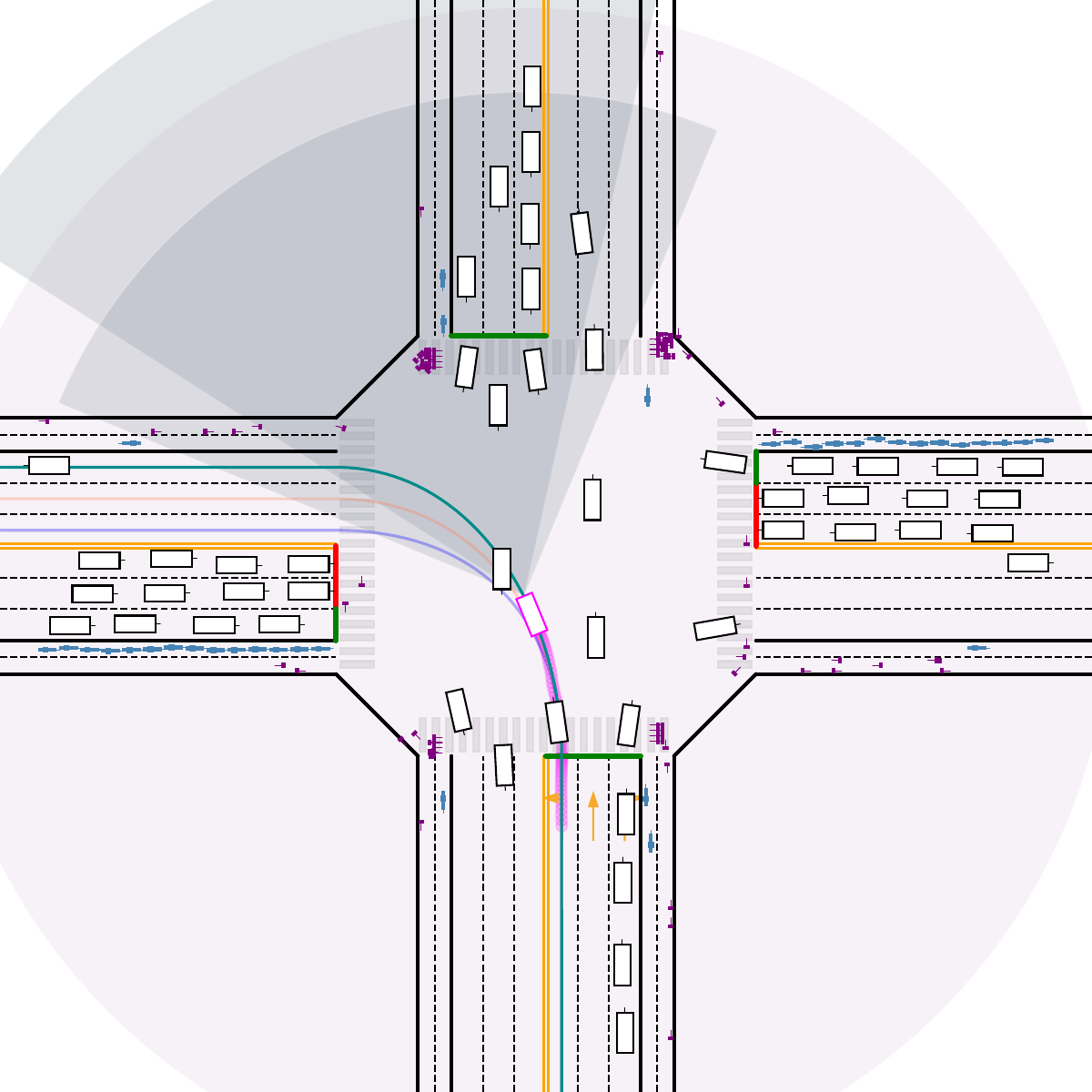}}
\subfloat[t=16.0s]{\includegraphics[width=0.25\linewidth]{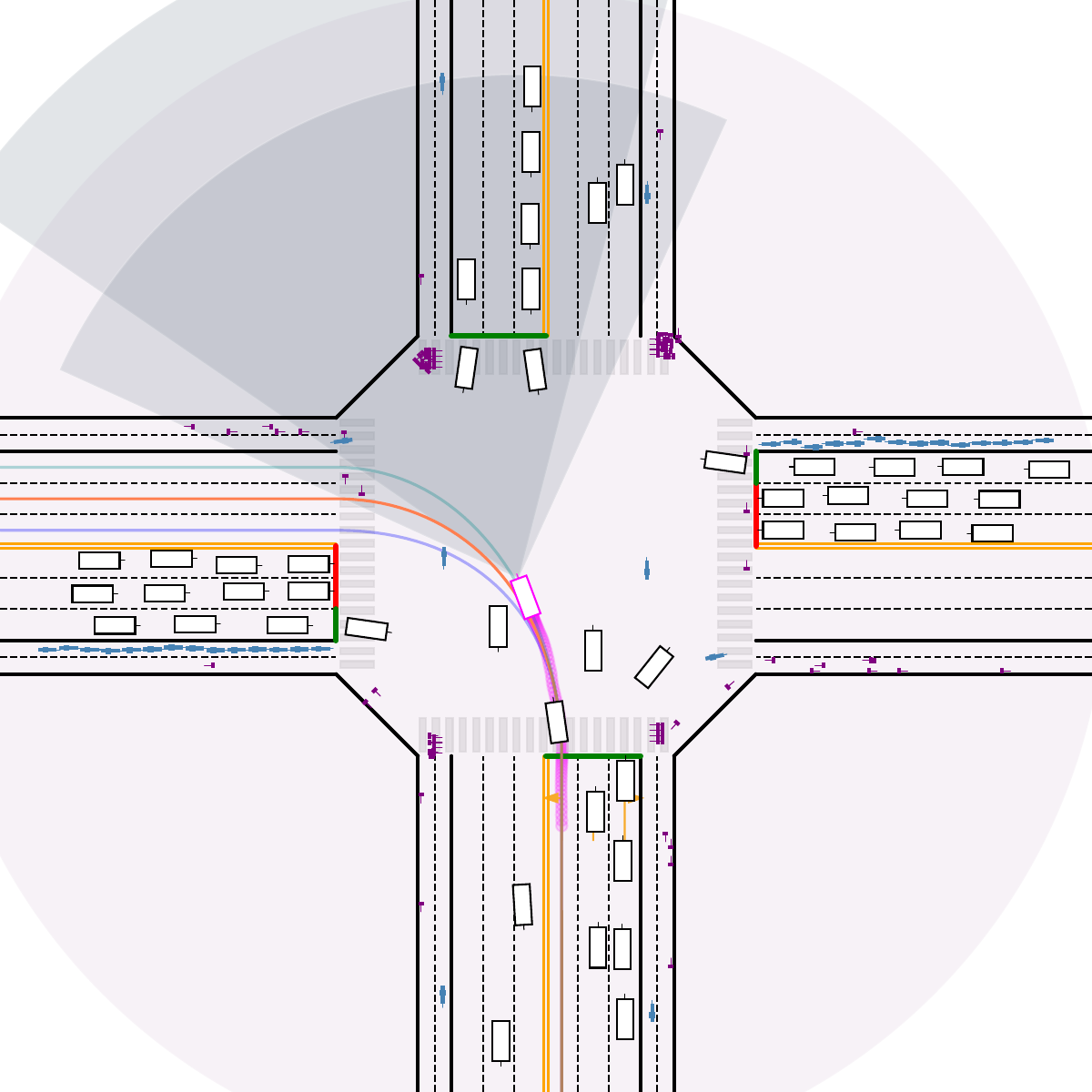}}
\subfloat[t=20.0s]{\includegraphics[width=0.25\linewidth]{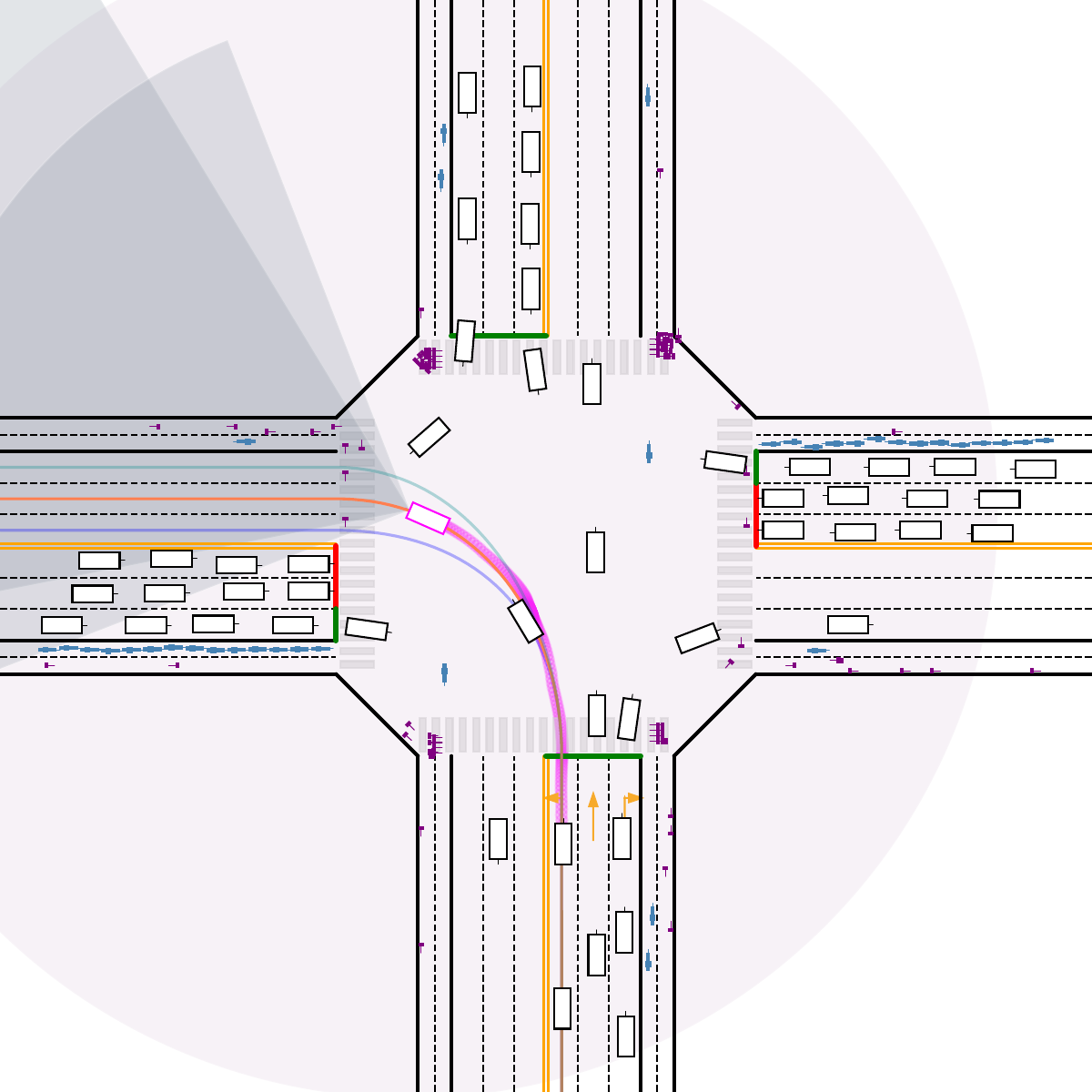}}
\\
\subfloat[t=22.5s]{\includegraphics[width=0.25\linewidth]{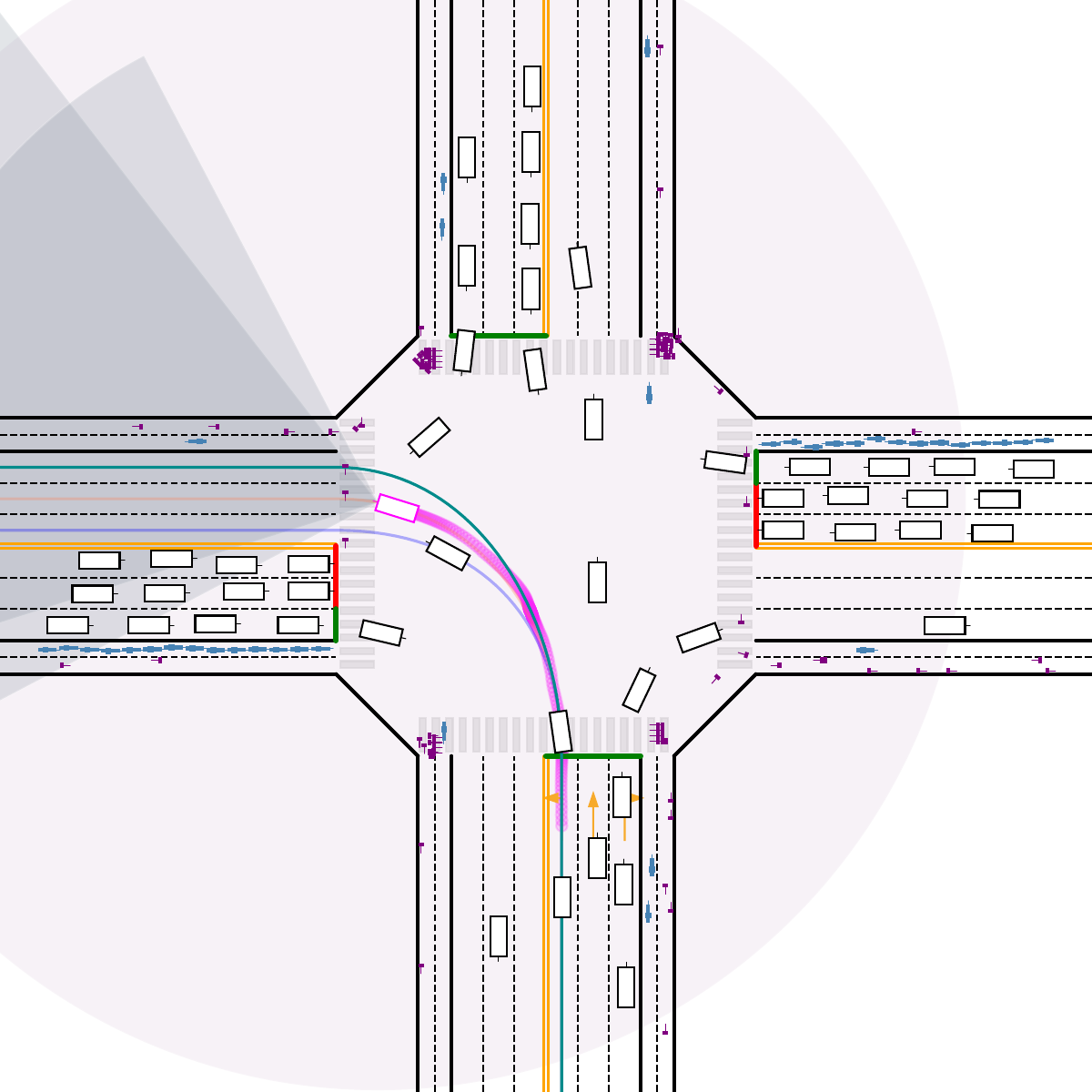}}
\subfloat[t=26.5s]{\includegraphics[width=0.25\linewidth]{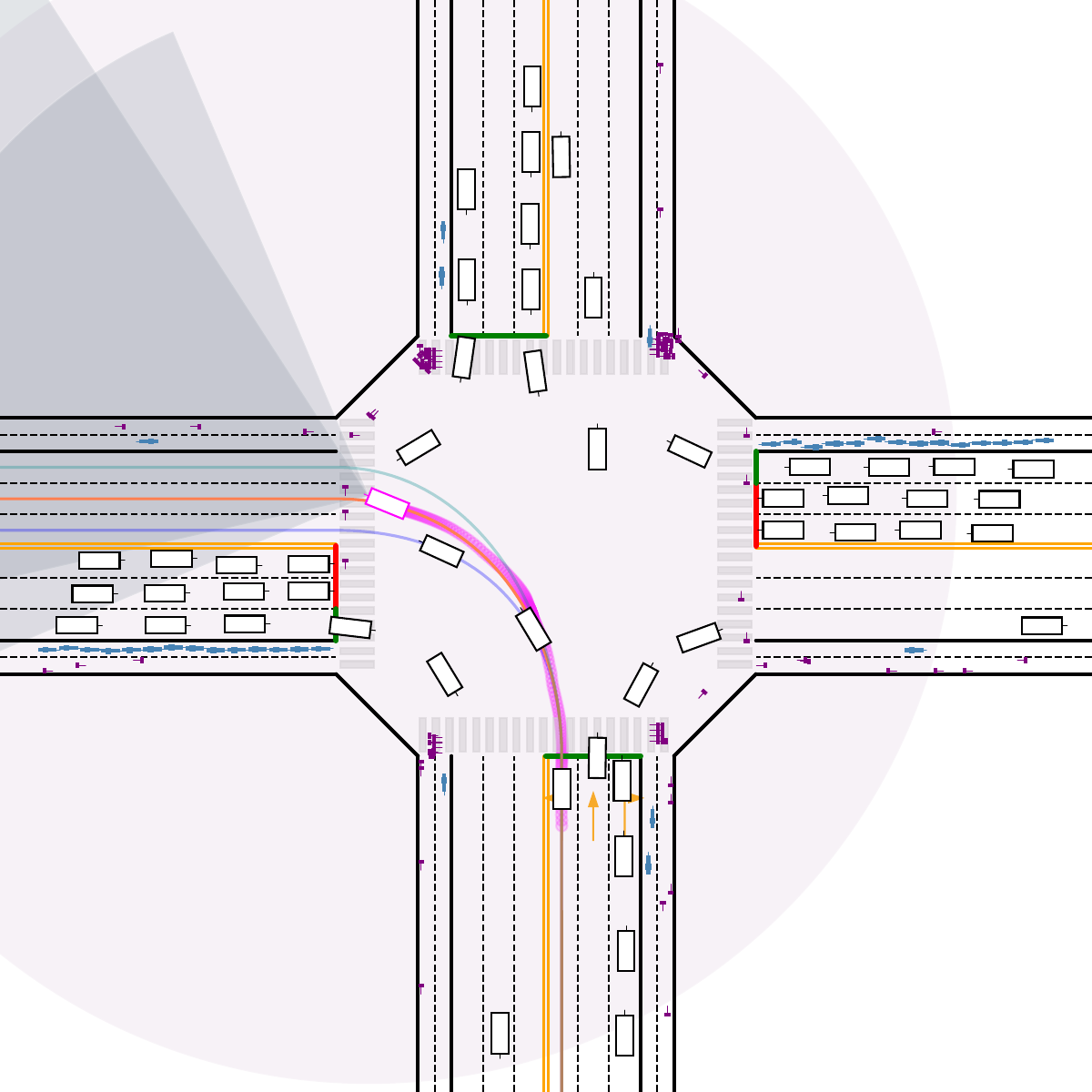}}
\subfloat[t=28.0s]{\includegraphics[width=0.25\linewidth]{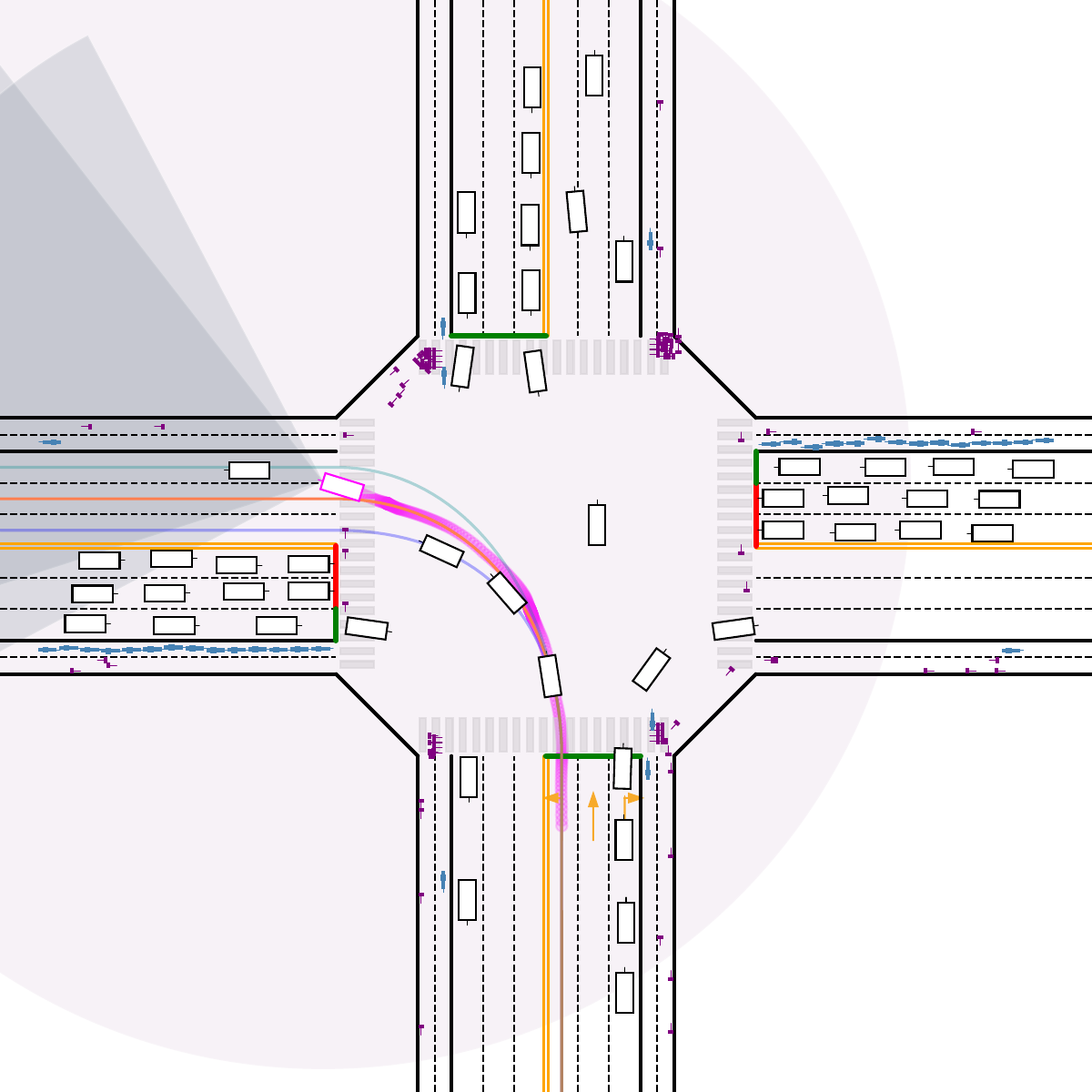}}
\subfloat[t=29.8s]{\includegraphics[width=0.25\linewidth]{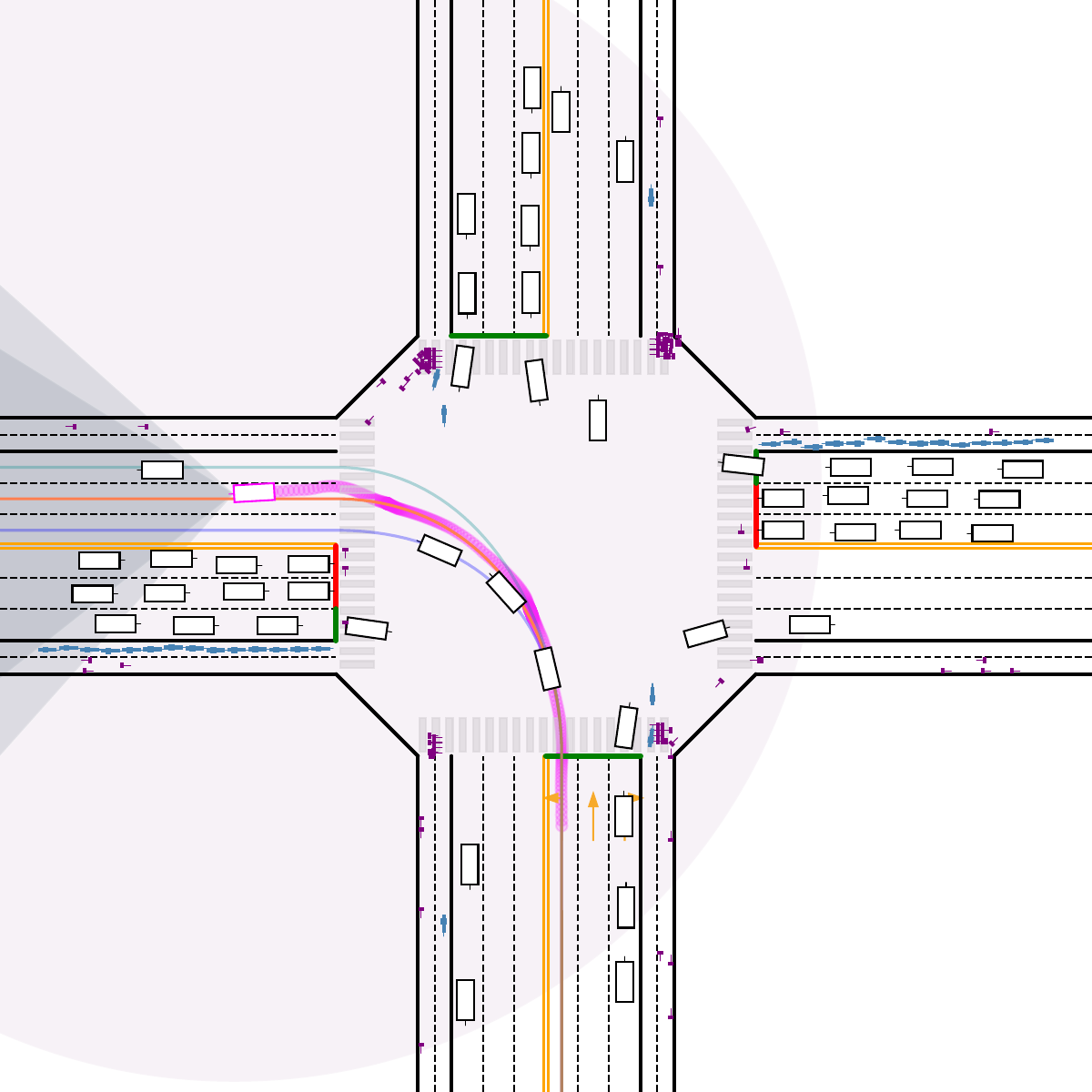}}
\caption{Trajectory visualization in the unprotected left turn task. The red box represents our ego vehicle controlled by the trained policy and the shaded sector indicates the perception range of different sensors.}
\label{fig.simu_traj}
\end{figure*}

\begin{figure}[!htb]
\captionsetup{justification =centering,
              singlelinecheck = false,labelsep=period, font=small}
\subfloat[Policy output: control commands of ego vehicle]{\label{fig:e1_control}\includegraphics[width=1.0\linewidth]{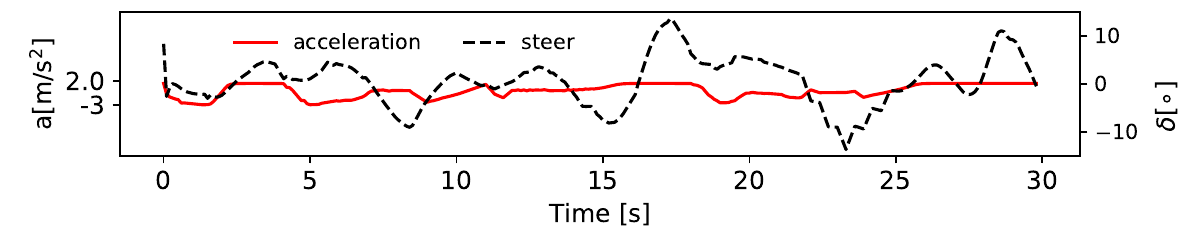}} \\
\subfloat[State of ego vehicle: speed and heading angle]{\label{fig:e1_angle}\includegraphics[width=1.0\linewidth]{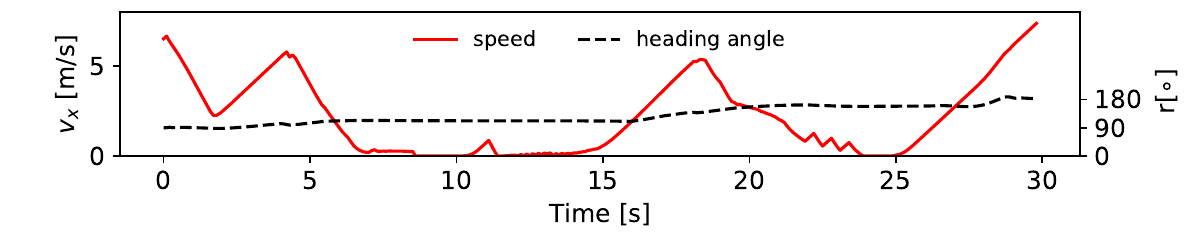}} \\
\subfloat[Value output: performance of candidate paths]{\label{fig:e1_angle}\includegraphics[width=1.0\linewidth]{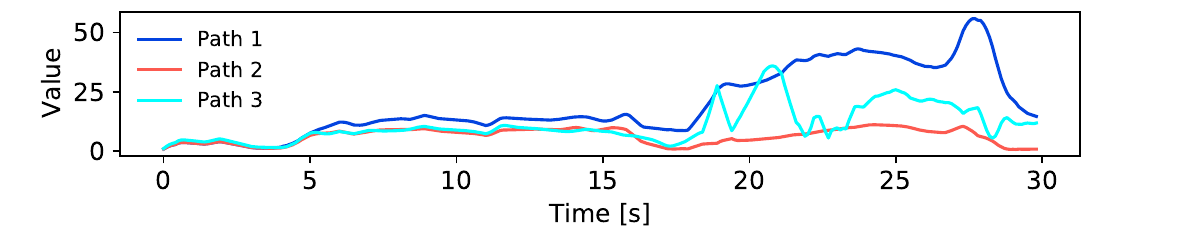}}
\caption{Control commands, state of ego vehicle and value of candidate paths}
\label{f:avg_speed}
\end{figure}

Next, we implement the trained networks of IDC with dynamic permutation state representation in the unprotected left turn task and visualize a typical driving process with the corresponding action, state and values of different paths in Fig.~\ref{fig.simu_traj} and Fig.~\ref{f:avg_speed}. This kind of left-turn is considered as one of the most difficult tasks of autonomous driving where the ego vehicle must choose the most appropriate target lane and deal with the moving surrounding participants.
The ego vehicle is initialized outside of the intersection with a high speed and it will decelerate gradually to wait for the straight passing of bicycles and vehicles from the opposite direction, as shown in Fig.~\ref{fig.simu_traj}(a) and (b). This waiting process can also be identified from the speed curve in Fig.~\ref{f:avg_speed}(b). Interestingly, the highlighted optimal path in Fig.~\ref{fig.simu_traj}(b) and the path values in Fig.~\ref{f:avg_speed}(c) show that the ego vehicle will choose the 3rd path during this process which brings more potential to bypass the straight vehicles. 
After that, from Fig.~\ref{fig.simu_traj}(c), (d) and Fig.~\ref{f:avg_speed}(c), we can see the ego vehicle tends to choose the 2nd path, i.e., the path with the lowest value, to track within this intersection as there exists fewer vehicles to arrive at this lane. When nearly reaching the sidewalk, the ego will decelerate to stop and yield to crossing pedestrians as shown in Fig.~\ref{fig.simu_traj}(e). 
Once the pedestrians walk further away, our ego vehicle will accelerate timely to bypass the pedestrians upward in Fig.~\ref{fig.simu_traj}(f)
$\sim$(h), by which a better passing efficiency is provably available. 
It can also be seen from Fig.~\ref{f:avg_speed}(a) and (b) that the ego vehicle can choose the optimal path by steering, avoid collision by decelerating and make a quick pass by accelerating, meanwhile the heading angle varies from $90^{\circ}$ to $180^{\circ}$ smoothly.

\subsection{Comparison of driving performance}
Here we analyze the driving performance of the learned policy riding at this intersection. In addition to IDC with two different state representations , we also introduce a rule-based baseline, in which the ego vehicle is controlled by the Krauss car-following and SL2015 lane-changing models of SUMO \cite{SUMO2018}. This human designed passing is on a first-come, first-go basis and pedestrians have the highest priority if encountering a conflict situation.
Referring to human experience, we design four indicators including comfort, time to pass, collisions and decision compliance to evaluate the driving performance.
Comfort is calculated by the mean root square of lateral and longitudinal acceleration;
Time to pass is evaluated by the average time used to pass the intersection, starting from entering the intersection at stop line; Collision means the ego vehicle hits its surrounding participants or rides out of this intersection and decision compliance shows times of breaking red light. Also, we record the computation time for calculating the control action for IDC, which could indicate the real time performance of applying the trained policy on vehicle computation platform.
For each method, 100 simulations will be conducted wherein the ego vehicle starts outside of the intersection with a random velocity. The maximum time length of each simulation is 180 seconds.
The results of driving performance are shown in Table \ref{table.comp}. For computation time, both of IDC algorithms can output the control commands within 10ms, which is promising to meet the real-time requirement of driving. Note that IDC with dynamic permutation state representation takes a little more time than IDC with fixed permutation method because of the introduction of encoding network $h(x, \phi)$. Rule-based method suffers from terrible comfort due to the neglect of vehicle dynamic model, and it is also more likely to stop and wait for other participants passing first, leading to a much longer passing time. By contrast, the two trained driving policies share almost the same pass time, but IDC with fixed permutation incurs more collisions, worse comfort, and more incompliant decisions. As we analyzed before, the fixed permutation state representation may cause the discontinuity of state concerning mixed traffic flows and poorly characterize the dynamics of driving environment, thus leading to the abruption or incorrectness of control actions.
\begin{table}[h]  
\caption{Comparison of driving performance}
\begin{tabular}{lccc}
\hline
                    &\makecell[c]{Dynamic\\ permutation}  &\makecell[c]{Fixed \\permutation}    & Rule-based \\ \hline
Computing time [ms]  & 7.24($\pm$1.36) & 5.96($\pm$0.60)  & -         \\
Comfort index       & 2.63         & 3.96          & 4.34          \\
Time to pass [s]    & 24.46($\pm$4.52) & 25.73($\pm$6.60)  & 66.18($\pm$15.74)          \\
Collisions          & 1            & 8             & 2            \\
Decision Compliance & 0            & 3             & 0            \\ \hline
\end{tabular}
\label{table.comp}
\end{table}

\subsection{Explanation of learned policy}
\begin{figure}[htbp]
\centering
\captionsetup[subfigure]{justification=centering}
\subfloat[Front wheel angle]{\label{fig.exp1.acc}\includegraphics[width=1.0\linewidth]{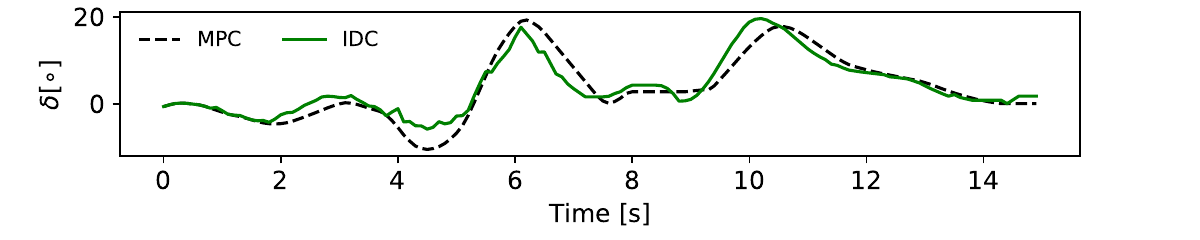}} \\
\subfloat[Acceleration]{\label{fig.exp1.front}\includegraphics[width=1.0\linewidth]{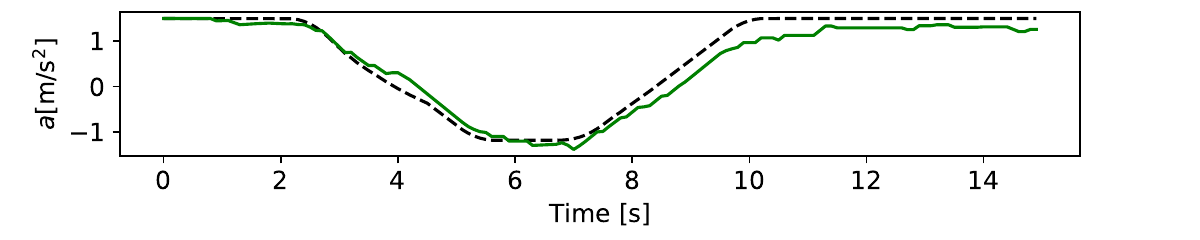}}
\caption{Control actions of MPC and IDC.}
\label{fig.exp1}
\end{figure}

We have demonstrated that IDC can improve the driving performance by combining with dynamic permutation state representation. Now we aim to identify what kinds of information the policy has learned. Actually, model predictive control (MPC) usually is adopted to solve the finite horizon constraint problem by utilizing the receding horizon optimization. It is a typical online optimization method and can deal with constraints explicitly\cite{garcia1989model}. Therefore, we
employ MPC to calculate the control action for the original problem  \eqref{eq.rl_problem_policy} at each observation $\SPACE{O}$ with the open-source solvers\cite{wachter2006implementation}. We choose one episode of left turn randomly and compare the control actions calculated by MPC and the output of the trained policy in Fig.~\ref{fig.exp1}.
Results show that the output actions (steer wheel and acceleration) have minor difference with the same input, indicating the policy indeed has learned to approximate the control effects of MPC, while the latter can be seen the optimal solution for original constraint optimal problem. 
However, there exists the obvious difference in computation time that our method can output the actions within 10ms while MPC will take an average time of 1000ms to perform that on this task of Fig.~\ref{fig.exp1}. 
To sum up, with dynamic permutation state representation, IDC is promising to approximate the exact solution of online optimization by training an optimal policy offline on the whole state space.

%% file: content/6Conclusion.tex
\section{Conclusion}
\label{sec:conclusion}

This paper focuses on the decision-making and control for signalized intersections with mixed traffic flows. 
To that end, we develop the dynamic permutation state representation in framework of integrated decision and control (IDC), which composes of an encoding function to construct driving states, a value function to choose the optimal path as well as a policy function to output the control command of ego vehicle. 
A constraint optimal problem is formulized to optimize these three functions, where the objective involves tracking performance within a finite horizon and the constraints aims to assure safety w.r.t. different participants and signal lights. 
Specially, the dynamic permutation state representation introduces this encoding function and summation operator to construct driving states from environmental observation, capable of dealing with different types and variant number of traffic participants. 
Finally, a complex urban intersection scenario is constructed to verify the effectiveness. 
Results indicate that dynamic permutation state representation can enhance the driving performance of IDC and realize intelligent and efficient passing under random traffic flows. About the future work, we will further improve the driving performance with more powerful encoding functions, for example, the state-of-the-art transformer network, which may extract more efficient features due to the superior representation ability.